\documentclass[letterpaper]{article} % DO NOT CHANGE THIS
\usepackage[preprint]{aaai2027}  % DO NOT CHANGE THIS
\usepackage[hyphens]{url}  % DO NOT CHANGE THIS
\usepackage{graphicx} % DO NOT CHANGE THIS
\usepackage{natbib}  % DO NOT CHANGE THIS AND DO NOT ADD ANY OPTIONS TO IT
\usepackage{caption} % DO NOT CHANGE THIS AND DO NOT ADD ANY OPTIONS TO IT
\usepackage{algorithm}
\usepackage{algorithmic}

\usepackage{newfloat}
\usepackage{listings}
\DeclareCaptionStyle{ruled}{labelfont=normalfont,labelsep=colon,strut=off} % DO NOT CHANGE THIS
\floatstyle{ruled}
\newfloat{listing}{tb}{lst}{}
\floatname{listing}{Listing}

\usepackage{booktabs}
\usepackage{amsmath}
\usepackage{amssymb}
\usepackage{mathtools}
\usepackage{amsthm}
\usepackage{threeparttable}
\usepackage[most]{tcolorbox}
\usepackage[dvipsnames]{xcolor}
\usepackage[table]{xcolor}
\usepackage{booktabs}
\usepackage{multirow}
\usepackage{makecell}
\usepackage{adjustbox}
\usepackage{textcomp}

\usepackage{minitoc}
\usepackage{etoc}

\DeclareUnicodeCharacter{3000}{\textcolor{red}{IAMAUNICODEWHITESPACE}}

\newtcolorbox{empheqboxed}{colback=Gray!10, 
 colframe=white,
 width=\linewidth,
 sharpish corners,
 top=1mm, %
 bottom=0pt,
 left=2pt,
 right=2pt
}

\newtcolorbox{remarkbox}{colback=Blue!5, 
 colframe=white,
 width=\linewidth,
 sharpish corners,
 top=1mm, %
 bottom=0pt,
 left=2pt,
 right=2pt
}

\theoremstyle{plain}
\newtheorem{theorem}{Theorem}

\theoremstyle{definition}

\theoremstyle{remark}
\newtheorem{remark}[theorem]{Remark}

\usepackage[textsize=tiny]{todonotes}
\usepackage{amsmath,amsfonts,bm}

\def\eqref#1{equation~\ref{#1}}
\def\1{\bm{1}}

\DeclareMathAlphabet{\mathsfit}{\encodingdefault}{\sfdefault}{m}{sl}
\SetMathAlphabet{\mathsfit}{bold}{\encodingdefault}{\sfdefault}{bx}{n}

\usepackage{hyperref} 
\usepackage[capitalize,noabbrev]{cleveref}

\title{Latent Actions from Factorized Transition Effects under Agent Ambiguity}

\affiliations {
\vspace{-20pt}
    \textbf{Heejeong Nam\footnote{Correspondence to: \texttt{heejeong\_nam@brown.edu}} \quad Chandradithya S Jonnalagadda \quad Harshit Aggarwal \quad Eric Xu \quad Randall Balestriero} \\ \vspace{5pt}
    \large Brown University\\ \vspace{10pt}
    \href{https://hazel-heejeong-nam.github.io/LAM/}{https://hazel-heejeong-nam.github.io/LAM/}\\
}

\begin{document}

\maketitle

\begin{abstract}

Latent Action Models (LAMs) learn action-like proxies from observation. However, in multi-object or distractor-rich scenes, observations contain not only agent motion but also distractors, camera dynamics, and background changes, making recovery of the underlying action intrinsically ambiguous without supervision. We argue that the appropriate unsupervised target is therefore not the true action itself, but a state-conditioned compositional summary of the transition effects present in the scene, enabling better action alignment and more effective utilization of action supervision.
To this end, we propose a two-stage framework. We first pretrain Observed Transition Factorization (OTF) to discover reusable local transition primitives using a compositional codebook. We then aggregate these primitives into compact state-conditioned latent actions, instantiated as OTF-LAM-Pixel under the standard inverse–forward dynamics framework and OTF-LAM-Dino, a decoder-free variant operating in a frozen DINOv2 representation space.
Experiments show that the learned transition primitives transfer across visual appearance and morphology, while the resulting latent actions exhibit substantially stronger action alignment than existing LAMs, make more effective use of downstream action supervision, and achieve competitive or superior downstream policy performance.
\end{abstract}

\section{Introduction}

World models~\cite{wm} for planning and control have largely been evaluated in settings where the controlled system and its action interface are predefined, including robotic manipulation and embodied-control benchmarks~\cite{hansen2024tdmpc,hafner2025mastering}. Their dynamics models predict future states conditioned on observed actions, providing a direct statistical cue for distinguishing action-dependent variation from autonomous scene dynamics, although this distinction need not be represented explicitly.
\begin{figure}[t]
\centering
\includegraphics[width=\linewidth]{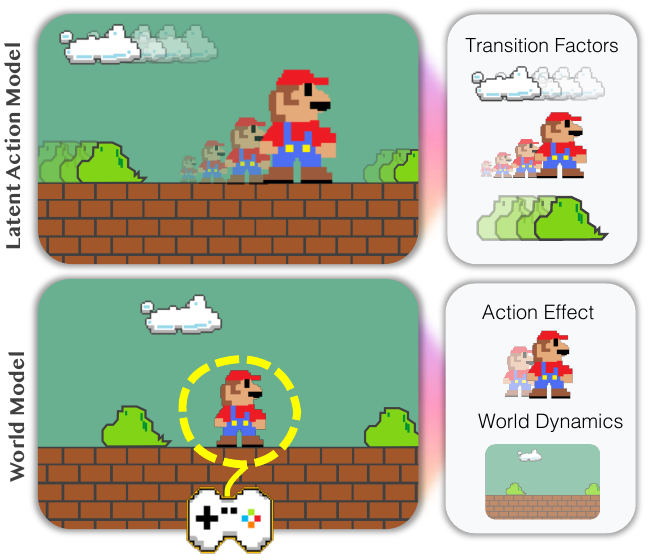}
\caption{
Motivation for agent ambiguity in observation-only latent action learning.
Unlike action-conditioned world models, which receive the actor and action as part of the input, LAMs infer action-like latents only from visual transitions.
}
\label{fig:teaser}
\end{figure}
Unlike conventional action-conditioned world models, Latent Action Models (LAMs)~\cite{lapo,lapa} learn action-like proxies directly from observation pair $(x_t,x_{t+1})$ without access to action labels or proprioceptive signals. 
This observation-only formulation fundamentally limits the recovery of true actions, because LAMs observe only their visual consequences, which may arise from agent motion, autonomous dynamics, camera motion, rendering, or other transition sources. While this distinction is often hidden in controlled environments where the agent dominates the scene~\cite{pmlr-v100-yu20a, robomimic2021}, real-world observations typically contain multiple transition sources simultaneously~\cite{laom, flam, segmentlam, oclam}. As illustrated in Figure~\ref{fig:teaser}, there is therefore no principled basis for deciding which transition effects should define the latent action representation without additional supervision or inductive biases.
Existing approaches typically address this ambiguity from an object-centric perspective, either by identifying the controllable object~\cite{segmentlam, oclam} or by learning separate latent actions for individual object-centric factors~\cite{lpwm, flam}. While effective in many settings, these approaches introduce a strong dependency on accurate object separation, even though objects are not always the right unit of abstraction, as transitions can manifest within, across, or independently of strict object boundaries. Independent of these practical limitations, the very need for such approaches reflects a broader reality: in unsupervised latent action learning, there is no canonical distinction between action-relevant and irrelevant transition effects. 
This leaves open a key question for us: how should multiple transition sources be represented under observation-only learning?
% 여기 아래 문장 살짝 verbose
To answer this question, we distinguish three levels in the causal pathway from actions to observations, as shown in Figure~\ref{fig:3step}. A true underlying action induces a physical or environment state transition, which in turn gives rise to the visual transition effects observed in image space. While the action itself is hidden, these transition effects are directly observable.
This perspective suggests that the appropriate unsupervised target for a LAM is a state-conditioned, compositional summary of the transition effects present in the scene. 
We therefore take a bottom-up approach to latent action learning under
ambiguity: rather than attempting to recover true actions directly from
observation-only transitions, we seek a structured intermediate
representation that organizes heterogeneous transition effects into reusable
components and can be effectively aligned with true action semantics through
downstream action supervision.
To realize this bottom-up approach, we separate latent action learning into two stages. We first pretrain an \textbf{Observed Transition Factorization (OTF)} module to discover reusable local visual transition primitives directly from observation. OTF learns a compositional VQ-style codebook that factorizes visual changes into sparse combinations of localized transition, minimizing reliance on environment- and embodiment-specific appearance while promoting transfer across visual carriers and morphologies.
Building upon this pretrained transition vocabulary, we aggregate identified transition primitives—capturing which primitives occur and where they are expressed—into compact, state-conditioned latent actions using a state-aware aggregator. We instantiate this idea as \textbf{OTF-LAM-Pixel} within the standard inverse–forward dynamics framework and as \textbf{OTF-LAM-Dino}, a decoder-free variant that predicts future states in a frozen DINOv2~\cite{dinov2} representation space.
Empirically, pretrained OTF primitives transfer across held-out visual carriers and cross-morphology shifts, indicating that they capture reusable
transition structure rather than environment- or embodiment-specific templates. Although LAPO, a representative LAM baseline with a monolithic latent
embedding, learns latent actions that are slightly easier to predict through behavioral cloning, OTF-LAM-Dino achieves substantially lower action-decoding
error under identical downstream training settings, indicating stronger alignment with the underlying environment actions. This improved action alignment translates into more effective use of action supervision and competitive or superior downstream policy performance relative to existing
LAM baselines. OTF-LAM-Dino consistently outperforms decoder-based OTF-LAM-Pixel, demonstrating the benefit of combining a reusable transition
vocabulary with a frozen pretrained visual representation.

\section{Problem Formulation}
\label{sec:comp}
 
Before introducing our method, we formalize the forward process that generates observations and the backward process through which a model learns from them. This distinction clarifies both why distractor ambiguity is intrinsic to observation-only learning and why the true underlying actions cannot, in general, be uniquely identified.

\subsection{The Forward Generative Chain}
\label{sec:forward}
\begin{figure}[t]
\centering
\includegraphics[width=\linewidth]{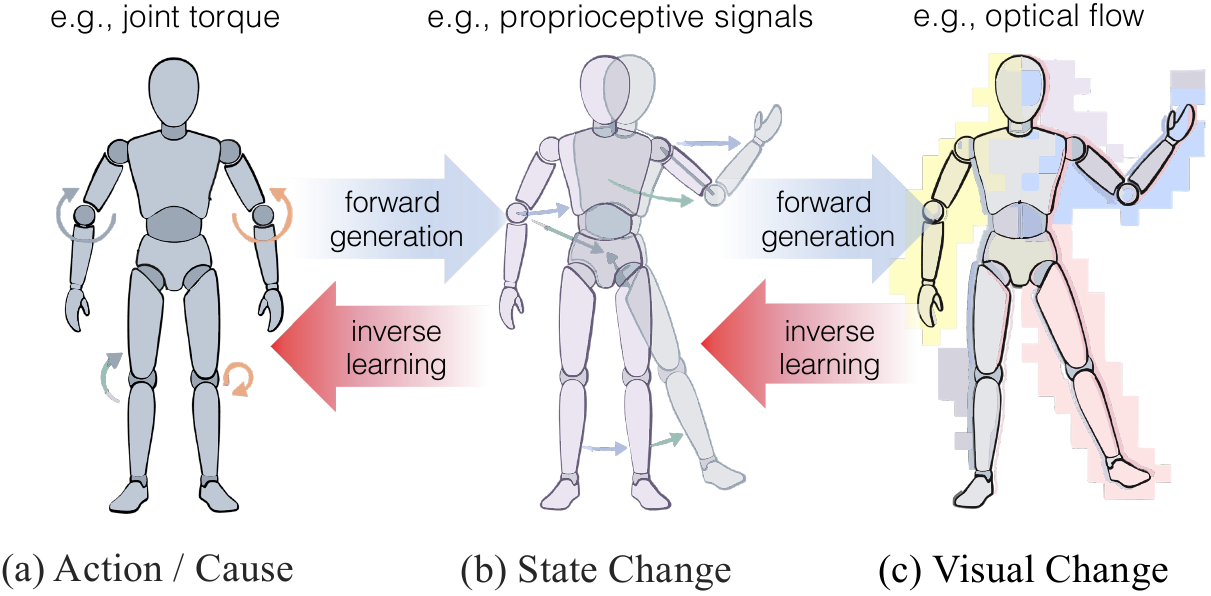}
\caption{
From causal generation to inverse inference. 
Blue arrows denote the causal pathway from action to visual change, while red arrows denote the inverse problem faced by observation-only latent action learning.
}
\label{fig:3step}
\end{figure}

We view a transition as a chain of three stages:
\[
\underbrace{a_t, \xi_t}_{\text{transition sources}}
\;\xrightarrow{F_{\mathcal{E}}}\;
\underbrace{\Delta s_t}_{\text{physical state change}}
\;\xrightarrow{R_{\mathcal{E}}}\;
\underbrace{\Delta x_t}_{\text{ transition effect}}
\]
as in Figure~\ref{fig:3step}.
Here, $a_t$ denotes the controlled action, while $\xi_t$ denotes uncontrolled transition sources such as camera motion, distractor motion, background dynamics, or passive environment changes.
The environment $\mathcal{E}$ includes the embodiment, dynamics, camera, and other environment-specific factors.
Let $F_{\mathcal{E}}$ be the state transition function and $R_{\mathcal{E}}$ be the observation or rendering map.
The physical state change may be written as
$$
\Delta s_t = F_{\mathcal{E}}(s_t,a_t,\xi_t),
$$
while the visual transition effect may be written as
$$
\Delta x_t = R_{\mathcal{E}}(s_t,\Delta s_t,\xi_t).
$$
Equivalently, the observed transition is generated by the composition
\begin{equation}
\label{eq:transition}
    \Delta x_t
=
R_{\mathcal{E}}
\bigl(
s_t,
F_{\mathcal{E}}(s_t,a_t,\xi_t),
\xi_t
\bigr).
\end{equation}

\subsection{The Backward Learning Chain}
\label{sec:backward}

Observation-only learning proceeds in the reverse direction of the generative process.
Given only a pair of consecutive observations $(x_t,x_{t+1})$, the learner first extracts a representation of the observed transition effect,
\begin{equation}
\label{eq:backward_effect}
    z_t
    =
    B_{\theta}(x_t,x_{t+1}),
\end{equation}
where $B_{\theta}$ denotes an observation-only transition encoder and $z_t^{act}$ summarizes the visual change $\Delta x_t$.
Conceptually, the backward learning chain is: 
\[
\underbrace{\Delta x_t}_{\text{transition effect}}
\;\xrightarrow{B_\theta}\;
\underbrace{z_t^{act}}_{\text{latent action}}
\;\xrightarrow{\text{supervision}}\;
\underbrace{\hat a_t}_{\text{action semantics}}
\]
The first step is learned from observation-only data, using both
$x_t$ and $x_{t+1}$.
However, reversing the rendering function $R_{\mathcal E}$ is generally
ill-posed.
Distinct physical state changes may induce the same visual transition
effect:
\begin{equation}
\label{eq:rendering_equivalence}
R_{\mathcal E}(s_t,\Delta s_t,\xi_t)
=
R_{\mathcal E}(s_t,\Delta s'_t,\xi_t),
\qquad
\Delta s_t \neq \Delta s'_t.
\end{equation}
Thus, the observed visual transition does not necessarily uniquely
determine the underlying physical state change.
Even if the correct physical state change were known, reversing the transition function $F_{\mathcal E}$ is still generally ill-posed.
Different combinations of controlled actions and uncontrolled transition sources may produce the same physical state change:
\begin{equation}
\label{eq:transition_equivalence}
F_{\mathcal E}(s_t,a_t,\xi_t)
=
F_{\mathcal E}(s_t,a'_t,\xi'_t),
\qquad
(a_t,\xi_t) \neq (a'_t,\xi'_t).
\end{equation}
Therefore, observation-only learning cannot in general recover a unique
inverse of $R_{\mathcal E}\circ F_{\mathcal E}$, nor can it determine
which component of the transition should be interpreted as the controlled
action when multiple causal explanations are observationally equivalent.
Appendix A provides additional discussion, including examples illustrating this ambiguity.

% (wait but the naturally rising question is, isn't other models doing it naturally?) 

\section{Pretraining}

Actions and state changes are often environment-specific.
Actions depend on the control interface, such as torques, joint commands,
or button inputs, while physical changes depend on the current state,
embodiment, contacts, and dynamics. In contrast, observed visual effects
$\Delta x_t$ provide a weaker but more transferable description of a
transition. They capture how change appears in observation space, including
local displacement, edge shifts, rotation-like residuals, contact
deformation, and background drift. For example,
\[
\texttt{Mario moves right}
\quad\text{and}\quad
\texttt{Kirby moves right}
\]
may involve different bodies, control interfaces, and dynamics, yet induce
similar rightward displacement patterns in observation space.
This motivates using observed transition effects as an intermediate
representation.

\subsection{Observed Transition Factorization (OTF)}
\label{sec:stage1}
\begin{figure}[t]
\centering
\includegraphics[width=\linewidth]{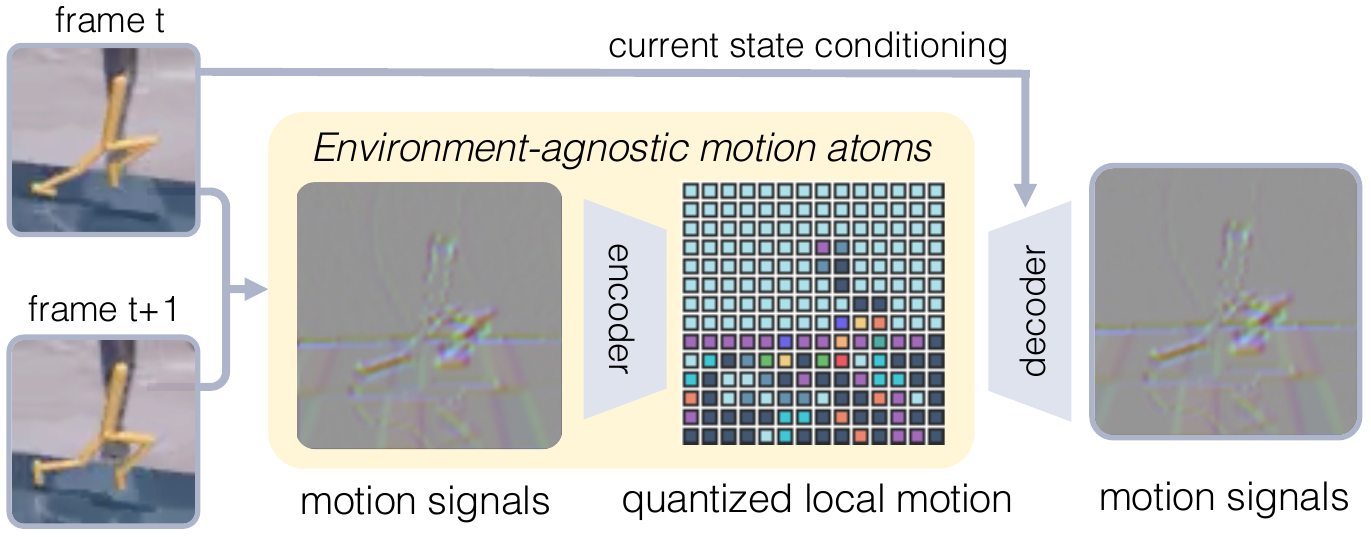}
\caption{
Observed-transition vocabulary learning with a VQ-VAE. Motion-centered transition inputs are patchified, quantized into a shared codebook of reusable observed-transition primitives, and decoded with reference-frame conditioning to reconstruct the observed-transition.
}
\label{fig:stage1}
\end{figure}
% 두괄식 설명해야함. 
We introduce a pretraining stage for tokenizing local visual changes. We call this module Observed Transition Factorization (OTF). It learns a shared codebook of $K$ reusable observed-transition primitives that capture recurring patterns such as local displacement, edge shifts, rotation-like residuals, contact deformation, background drift, and articulated motion. The key objective is to tokenize motion while minimizing environment- and embodiment-specific information.

\paragraph{Motion space.}
We define the motion space $o_t$ as a frame-level transition signal over stride $\tau$. To reduce dependence on RGB appearance and object identity, we apply a gradient transform to both frames and subtract the transformed current frame from the transformed future frame.

\begin{figure*}[t!]
\centering
\includegraphics[width=\textwidth]{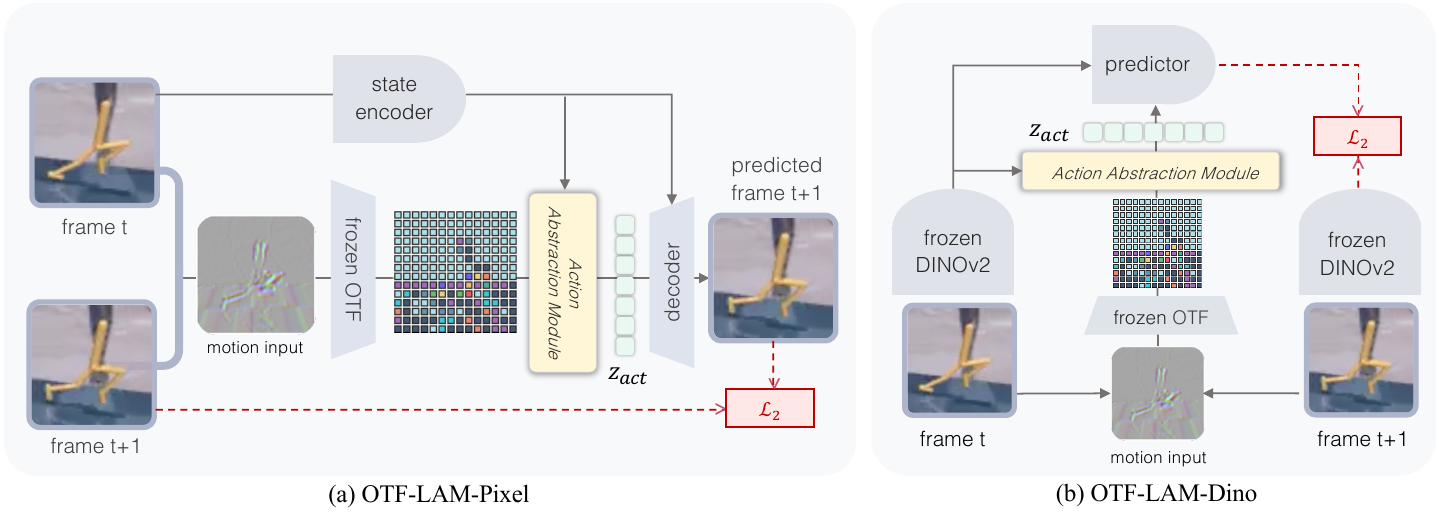}
\caption{Latent-action learning on top of the pretrained observed-transition vocabulary. OTF-LAM-Pixel predicts the future frame in pixel space, whereas OTF-LAM-DINO predicts the future representation in a frozen DINOv2 feature space.}
\label{fig:architecture}
\end{figure*}

\paragraph{Patchwise vector-quantized encoding.}
Given $o_t \in \mathbb{R}^{C \times H \times W}$, we divide it into $P$ non-overlapping spatial patches and encode each patch with a shallow MLP into $f_{t,i} \in \mathbb{R}^D$.
Each token is assigned to its nearest codebook entry by top-1 nearest-neighbor quantization:
\begin{equation}
\label{eq:quantization}
k^*_{t,i}
=
\arg\min_{k \in \{1,\dots,K\}}
\|f_{t,i}-c^{(k)}\|_2^2,
\qquad
q_{t,i}=c^{(k^*_{t,i})}.
\end{equation}
This quantization discretizes local observed effects, rather than dense optical-flow vectors.
From the patch assignments, we derive an occupancy map $M_t^{(k)}$ and an activation strength $w_t^{(k)}$ for each code $k$:
\begin{equation}
\label{eq:stage1return}
E_t
=
\left\{
e_{t,k}
\right\}_{k=1}^{K},
\qquad
e_{t,k}
=
\bigl(c^{(k)}, M_t^{(k)}, w_t^{(k)}\bigr).
\end{equation}
% Thus, $E_t$ records which primitives are active, where they occur, and how strongly they are expressed.

\paragraph{Frame-conditioned decoding.}
Although $o_t$ suppresses object-level semantics, local tokens may still retain carrier-specific cues such as texture, contrast, geometry, and occlusion. Patch size controls this trade-off: smaller patches reduce appearance leakage but provide less motion context, whereas larger patches capture richer motion evidence but may mix multiple transition sources and become tied to local appearance. We discuss this trade-off in Appendix B.2.
To reduce this burden on the quantized pathway, we condition the decoder on the current frame. The frame is encoded into a feature map aligned with the $H_p \times W_p$ motion grid and concatenated locally with the reconstructed factor map. The frame branch supplies appearance and spatial support, allowing the quantized pathway to represent only the complementary transition effect. This not only simplifies decoding, but also shapes the encoder toward motion-focused representations. A formal description is provided in Appendix B.3.

\paragraph{Training.}
As illustrated in Figure~\ref{fig:stage1}, we train OTF as a VQ-VAE to
reconstruct the motion-space target $o_t$ rather than the future RGB frame.
Each active transition factor is mapped to an embedding by a shared network
$\rho_\eta$ and placed back onto the patch grid according to its occupancy
map:
\[
H_t(u,v)
=
\sum_{k=1}^{K}
M_t^{(k)}(u,v)\,
\rho_\eta(e_{t,k}).
\]
The current frame is encoded into a feature map aligned with the
$H_p\times W_p$ motion grid and concatenated locally with the reconstructed
factor map. This reduces the information that must pass through the
quantized motion primitives by supplying appearance and spatial support directly to
the decoder. Finally, we predict $\hat{o}_t=D_\eta(H_t,x_t)$.
OTF is trained with a motion reconstruction objective, standard VQ losses, and an orthogonality regularizer:
\[
\begin{aligned}
\mathcal{L}_{\mathrm{vocab}}
&=
\mathcal{L}_{\mathrm{rec}}
+
\lambda_{\mathrm{1}}\mathcal{L}_{\mathrm{code}}
+
\lambda_{\mathrm{2}}\mathcal{L}_{\mathrm{commit}}
+
\lambda_{\mathrm{3}}\mathcal{L}_{\mathrm{orth}},
\end{aligned}
\]
where
\begin{align}
\mathcal{L}_{\mathrm{rec}}
&= \|\hat{o}_t - o_t\|_2^2, \quad
\\
\mathcal{L}_{\mathrm{code}}
&=
\sum_{i=1}^{P}
\left\|
\operatorname{sg}(f_{t,i}) - q_{t,i}
\right\|_2^2 \\
\mathcal{L}_{\mathrm{commit}}
&=
\sum_{i=1}^{P}
\left\|
f_{t,i} - \operatorname{sg}(q_{t,i})
\right\|_2^2, \\
\mathcal{L}_{\mathrm{orth}}
&=
\sum_{\substack{k,k' \in \mathcal{I}_t \\ k \neq k'}}
\left(
\hat{\mathbf{c}}^{(k)\top}
\hat{\mathbf{c}}^{(k')}
\right)^2.
\end{align}
More details are provided in Appendix B.4.

\remark[\textbf{Local Quantization vs. Discrete Latent Actions.}]{
Recent work~\cite{garrido} suggests that using vector quantization as the final latent action bottleneck can be too restrictive for complex in-the-wild videos. 
OTF uses quantization differently. The codebook represents local observed-transition primitives, while the final action-like latent is obtained by aggregating these primitives into a continuous representation. 
Thus, the discrete bottleneck is used to structure reusable local effects, not to represent the full action as a single code.
}

% \begin{remarkbox}
% \looseness=-1 \textbf{Remark 1. Reusability under Agent Ambiguity.}
% In this paper, reusability is not only a training convenience, but also a way to delay agent attribution under ambiguity.
% When multiple transition sources are present, directly compressing the transition into a single latent action can entangle agent-induced changes with distractors.
% A reusable vocabulary of local observed-transition primitives can describe motion effects under new distractors, morphologies, or distribution shifts before the relevant factors are aggregated into an action-like latent.
% \end{remarkbox}

% scaling up?

\section{Learning OTF-LAM}
\label{sec:stage2}

The frozen OTF encoder learns a reusable vocabulary of local observed-transition primitives. However, local transitions alone are insufficient for latent action learning, since the same primitive may arise from different causes depending on the scene. We therefore introduce an action abstraction module, to aggregate the transition factors in the context of the current observation to obtain a compact latent action representation. We then instantiate this idea with two variants: OTF-LAM-Pixel and OTF-LAM-DINO.

% why ours is better than object centric learning?

\subsection{Action Abstraction Module}
\label{sec:action_abstraction}

For action abstraction, we first construct a set of state-aware transition
tokens
\[
u_{t,i}
=
\phi_\theta
\left(
[s_{t,i},c_{t,i},p_i]
\right),
\qquad i=1,\ldots,N,
\]
where \(s_{t,i}\) denotes the state feature extracted from the current
observation at patch \(i\), \(c_{t,i}\) is the corresponding OTF transition
code, \(p_i\) is a positional embedding, and \(N\) is the number
of spatial locations. We denote the resulting token set by
\[
U_t=\{u_{t,i}\}_{i=1}^{N}.
\]
For aggregation, single global pooling operation of $U_t$ would compress all state-aware transition
tokens into a single representation using a fixed aggregation rule. Instead,
we introduce \(L\) learnable action queries $Q=\{q_\ell\}_{\ell=1}^{L}$, which allow the aggregation to
adapt to the input by collecting different subsets of transition tokens
before producing the final latent action.
Each query reads from the full set of state-aware transition tokens through
cross-attention
\[
H_t
=
Q+
\operatorname{CrossAttn}_\theta
\left(
Q,U_t,U_t
\right),
\]
making it input-dependent.
The resulting action tokens are further refined through self-attention,
allowing information aggregated by different queries to be integrated, then projected to the final latent action:
\[
z_t^{\mathrm{act}}
=
\phi_{\mathrm{out}}
\left(
\operatorname{Flatten}
\left(
\operatorname{Transformer}_\theta(H_t)
\right)
\right).
\]

\subsection{OTF-LAM-Pixel}

OTF-LAM-Pixel learns latent actions by predicting future observations in pixel
space. The current observation is first encoded by a learned state encoder,
whose patch-level features are used both by the Action Abstraction Module to
produce the latent action and by a pixel-space forward model to preserve the
current scene context. Conditioned on the latent action, the forward model
predicts the future observation using residual prediction by default.
The model is trained with a next-frame reconstruction objective,
\[
\mathcal{L}_{\mathrm{Pixel}}
=
\left\|
\hat{x}_{t+\tau}-x_{t+\tau}
\right\|_2^2.
\]
The observed-transition vocabulary remains frozen throughout training, and only
the state encoder, Action Abstraction Module, and forward model are optimized.
Architectural details are provided in Appendix C.2.

\subsection{OTF-LAM-DINO}

OTF-LAM-DINO replaces pixel prediction with representation prediction in a
frozen DINOv2~\cite{dinov2} feature space. Instead of a learned pixel encoder,
both the current and future observations are encoded by a frozen DINOv2
backbone. The current DINO patch features are used by the Action Abstraction
Module together with the frozen observed-transition vocabulary, while a
representation predictor estimates the future DINO representation conditioned
on the latent action.
The model is trained by matching the predicted and target DINO
representations,
\[
\mathcal{L}_{\mathrm{DINO}}
=
\left\|
\hat{S}_{t+\tau}-S_{t+\tau}
\right\|_2^2.
\]
Both the observed-transition vocabulary and the DINO encoder remain frozen,
leaving only the Action Abstraction Module and representation predictor to be
learned. Architectural details are provided in
Appendix C.3.

\section{Downstream Policy Learning}
\begin{figure}[t]
\centering
\includegraphics[width=\linewidth]{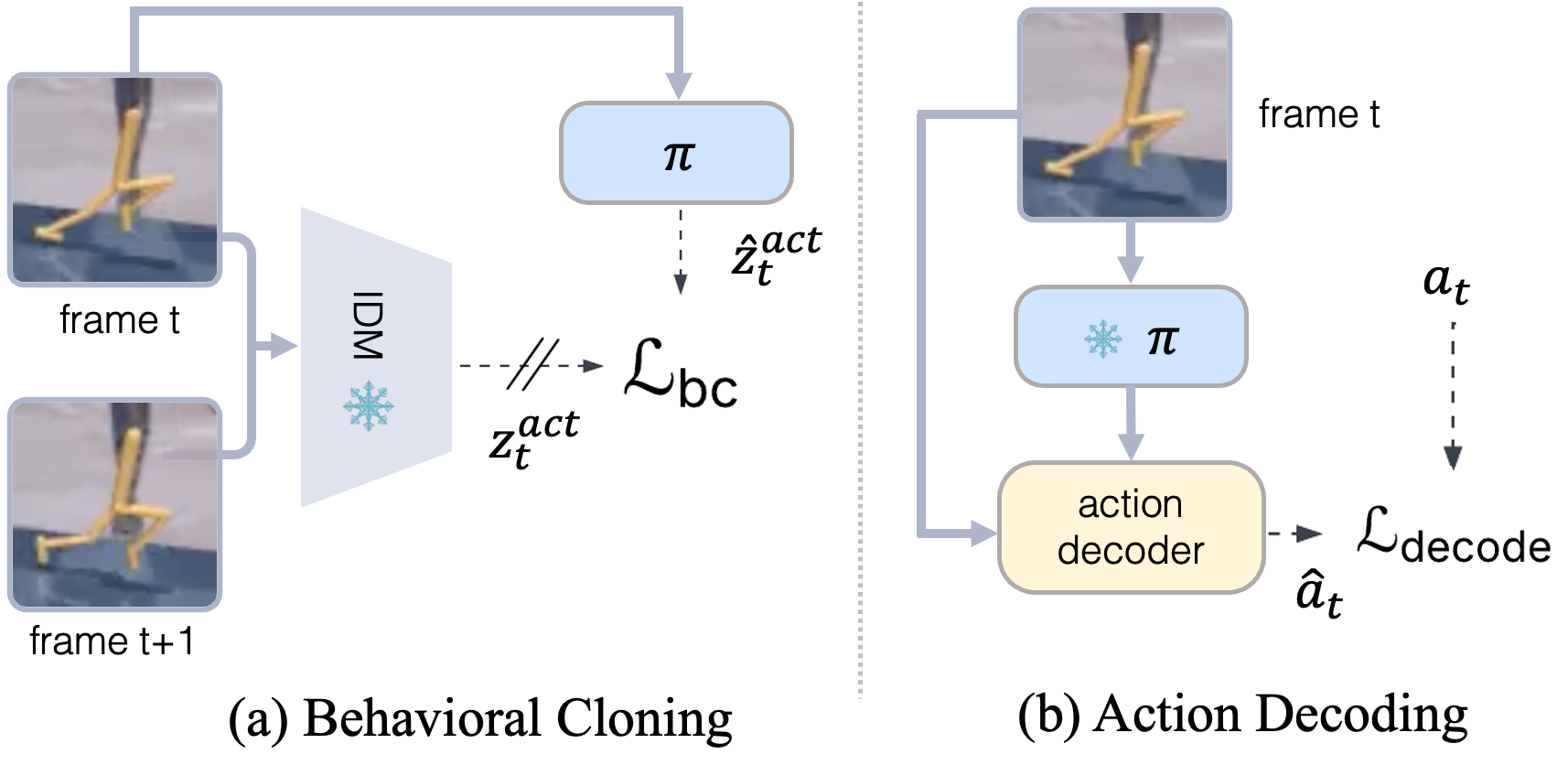}
\caption{
Policy Training. The latent motion space is distilled into a policy via behavioral cloning. An action decoder then maps these primitives to ground-truth actions $a_t$.
}
\label{fig:policy}
\end{figure}
A common downstream application of latent action models is policy learning, where
the quality of the learned latent actions is reflected in their ability to
support action prediction. The final step is therefore to connect the learned
latent action space back to real actions. Since the visual transition alone is
not invertible to the true action, we use a small amount of action supervision
to resolve this inverse ambiguity.
As described in Figure~\ref{fig:policy}, we first distill the learned latent
action space into a policy $\pi(z_t^{\mathrm{act}}\mid x_t)$ via behavioral
cloning. We then train an action decoder on a small action-labeled dataset to
map the predicted latent action to the true environment action. The decoder
also receives the current observation $x_t$, allowing it to associate latent
motion with the relevant object or state context. Unless otherwise noted, all methods, including the baselines, follow the same downstream protocol, with action decoders trained using 32 action-labeled trajectories.
\section{Experiments}

\paragraph{Datasets.}
We primarily evaluate on the Distracting Control Suite
(DCS)~\citep{dcs}, using \texttt{cheetah-run} and \texttt{walker-run}.
Trajectories are collected using pretrained expert policies from
\citet{laom}. For each environment, we collect 2,000 trajectories
(${\sim}2$M transitions), using 1,000 for latent action learning and the rest
for downstream training and evaluation.

We additionally use Moving MNIST~\citep{movingmnist} as a controlled benchmark
for zero-shot transfer of the observed-transition vocabulary. Digit identity
serves as the visual carrier, while transition primitives are shared across
classes. We train on 3,000 sequences with digits $\{0,1,2,3,4\}$ and evaluate
on 300 sequences with held-out digits $\{5,6,7,8,9\}$.

\paragraph{Baselines.}
We compare OTF-LAM against three existing latent action models, including
LAPO~\cite{lapo} (128/256 dimension latent action each), FLAM~\cite{flam} (4/8 slots each), and HiLAM~\cite{hilam}, which represent
monolithic, factorized, and hierarchical latent action learning,
respectively. We additionally evaluate two OTF-LAM-specific ablations,
Motion-LAPO and F-LAPO, to isolate the contributions of explicit motion
processing and patchwise latent quantization. Implementation details of all
baselines are provided in Appendix E.

\subsection{Transferability of Learned Motion Vocabulary}
\label{sec:learned_vocab}

We evaluate whether the observed-transition vocabulary learned by the pretrained OTF tokenizer captures reusable motion primitives that transfer across visual carriers and embodied morphologies. To this end, we train the vocabulary on \texttt{walker-run} and apply it zero-shot to \texttt{cheetah-run}, keeping the codebook fixed. As a controlled diagnostic, we additionally evaluate on Moving MNIST~\cite{movingmnist}, training on digits \texttt{\{0,1,2,3,4\}} and transferring zero-shot to held-out digits \texttt{\{5,6,7,8,9\}}. We compare against a monolithic VQ-VAE that represents each transition with a single global latent bottleneck under a comparable latent budget.

% We also study different motion inputs for vocabulary learning, comparing grayscale frames with a Sobel transform against our default gradient-based transform. Details of the motion input construction are provided in Appendix~\ref{app:motion}.

\begin{figure}[t]
\centering
\includegraphics[width=\linewidth]{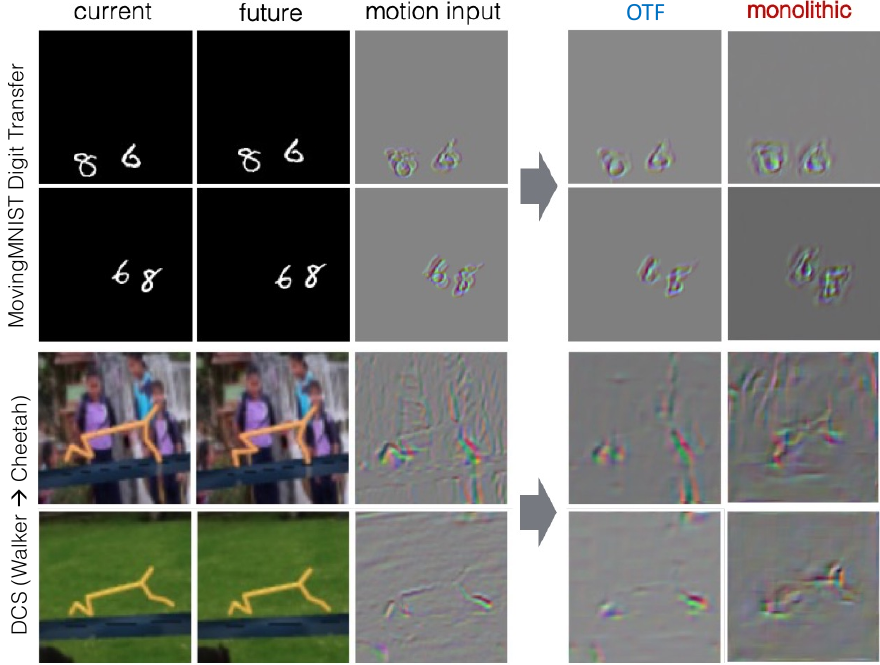}
\caption{
Zero-shot transfer of the learned observed-transition vocabulary. Comparison of motion reconstruction under cross-morphology transfer (DCS) and cross-carrier transfer (Moving MNIST).
}
\label{fig:dcs_transfer}
\end{figure}

\paragraph{Results on transferability.}

\begin{table}[t]
\centering
\label{tab:transfer}

\tabcolsep=0.17cm
\renewcommand{\arraystretch}{0.8}
\small

\begin{tabular}{@{}llccc@{}}
\toprule
\multirow{2}{*}{\textbf{Method}}
& \multirow{2}{*}{\textbf{Motion Input}}
& \multicolumn{2}{c}{\textbf{MSE}}
& \multirow{2}{*}{\textbf{Drop (\%)}} \\
\cmidrule(lr){3-4}
&
& Train
& Transfer
& \\
\midrule
\multirow{2}{*}{OTF}
& 1st-order & 0.037 & 0.055 & 49.34\% \\
&  2nd-order & 0.142 & 0.171 & 20.49\% \\
\midrule
\multirow{2}{*}{Monolithic}
& 1st-order  & 0.048 & 0.080
& 64.78\% {\color{red}($\uparrow$15.4)} \\
& 2nd-order & 0.172 & 0.271
& 58.18\% {\color{red}($\uparrow$37.7)} \\
\bottomrule
\end{tabular}
\caption{
Quantitative evaluation of cross-morphology zero-shot transfer degradation in DCS under different motion inputs (1st-/2nd-order temporal differences). The 2nd-order input emphasizes motion changes.
}
\end{table}

Figure~\ref{fig:dcs_transfer} shows qualitative transfer under embodiment and visual-carrier shifts, respectively (larger visualizations are provided in the Appendix D). Since consecutive frames are often visually similar, simply copying the reference frame can appear plausible despite failing to capture the true transition. The monolithic VQ-VAE frequently exhibits this failure mode under transfer, producing reconstructions that remain closer to the reference frame than to the target future frame. In contrast, OTF more faithfully preserves the target motion, suggesting that its factorized codebook captures reusable observed-transition primitives rather than source-specific appearance.
Quantitatively, we measure the relative transfer degradation
\[
\mathrm{Drop}
=
\frac{
\mathcal{E}_{\mathrm{target}}-\mathcal{E}_{\mathrm{source}}
}{
\mathcal{E}_{\mathrm{source}}
},
\]
where $\mathcal{E}_{\mathrm{source}}$ and $\mathcal{E}_{\mathrm{target}}$ denote the reconstruction errors on the source and transfer domains, respectively. Lower values indicate better transferability.
As shown in Table~\ref{tab:transfer}, OTF consistently exhibits substantially smaller degradation than the monolithic VQ-VAE across both motion inputs. These results indicate that OTF learns motion primitives that transfer more effectively across embodiments and visual carriers.

\subsection{Action Alignment of Learned Latent Actions}

\begin{figure}[t]
\centering
\includegraphics[width=\linewidth]{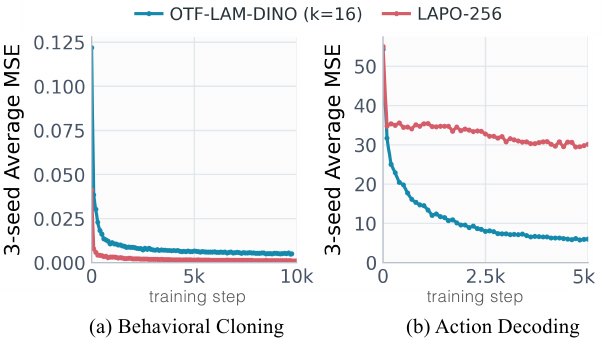}
\caption{
Behavioral cloning and action decoder convergence under same latent action capacity, comparing OTF-LAM-DINO and LAPO-256. 
}
\label{fig:converge}
\end{figure}

To understand the quality of the learned latent action space, we compare
behavioral cloning and action decoding under identical training settings.
Both OTF-LAM-Dino and LAPO-256 use a 256-dimensional latent action space and share
the same policy architecture, action decoder, optimization procedure, and
amount of action supervision.
Figure~\ref{fig:converge} (a) shows the behavioral cloning loss.
Across all supervision budgets, LAPO achieves slightly lower prediction error,
indicating that its latent actions are marginally easier to predict from
observations.
However, this trend reverses when decoding latent actions back to the
environment actions.
As shown in Figure~\ref{fig:converge} (b), OTF-LAM-Dino consistently converges
substantially faster and reaches much lower action-decoding error than LAPO,
regardless of the amount of supervision.
This contrast reveals that \emph{latent predictability} and
\emph{action alignment} are distinct properties.
A latent representation that is easier to imitate is not necessarily easier to
decode into the underlying control signal.
Instead, OTF-LAM-Dino learns latent actions that are considerably better
aligned with environment actions, making them substantially more
sample-efficient to supervise and more effective as an interface for downstream
policy learning.

\subsection{Policy Learning Results Against Baselines}

\begin{figure}[t]
\centering
\includegraphics[width=\linewidth]{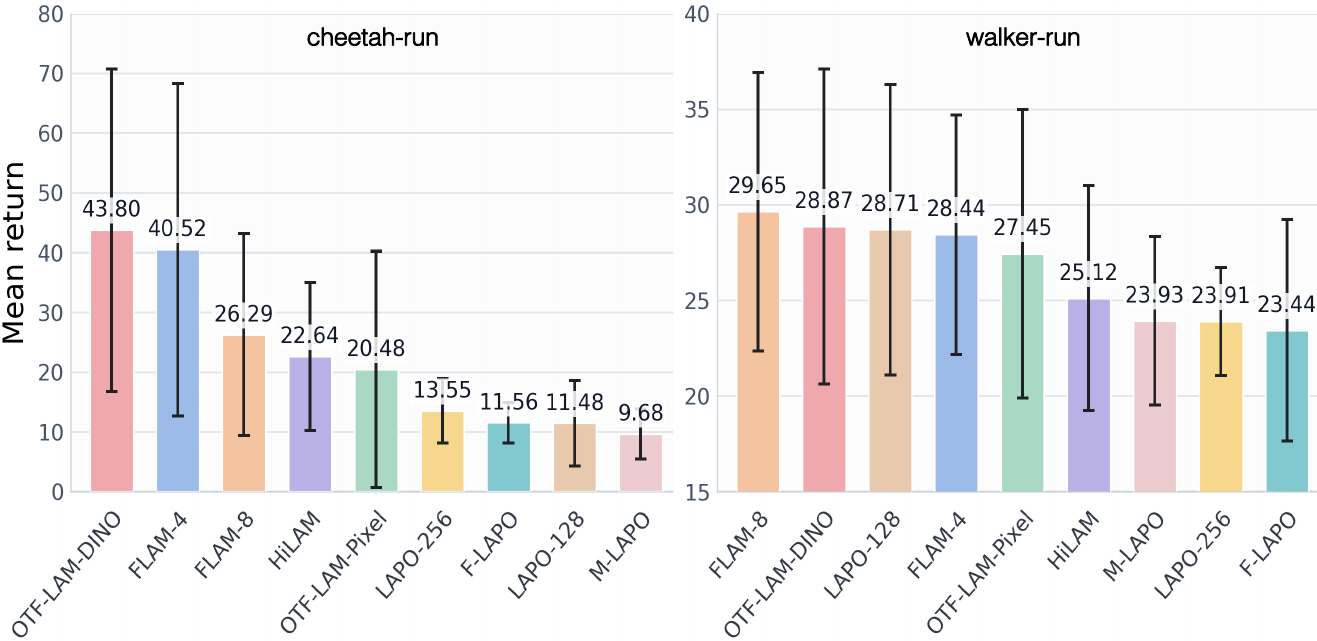}
\caption{
Bar chart comparing the mean return of LAMs on the DCS downstream policy evaluation task. Error bar is computed as the RMS of standard deviations.
}
\label{fig:combined_result}
\end{figure}

Figure~\ref{fig:combined_result} reports downstream policy performance on DCS.
For each seed, the policy is evaluated over $10$ trajectories, and each model is trained with three random seeds.
For OTF-LAM and OTF-LAM-Dino, we report the best-performing vocabulary size $K$.
The results highlight three observations.
First, OTF-LAM-Dino consistently outperforms decoder-based OTF-LAM in both environments.
Second, both OTF variants remain competitive with, and in the case of OTF-LAM-Dino outperform, existing latent action models despite relying on a reusable observed-transition vocabulary.
Third, methods with explicit scene-level factorization, such as FLAM, remain highly competitive, suggesting that compositional scene representations continue to provide benefits when such supervision is available.
The performance gain of OTF-LAM-Dino over OTF-LAM is likely attributable to the prediction target rather than the visual encoder itself.
Instead of reconstructing future pixels, OTF-LAM-Dino predicts future states in a frozen DINO representation, which suppresses many pixel-level nuisance factors while preserving control-relevant information.
This encourages the latent action model to focus on transition dynamics instead of appearance reconstruction.
Overall, these results suggest that OTF provides an effective latent action interface for downstream policy learning, while remaining compatible with both learned and frozen state representations.

\paragraph{Full Results.}
Appendix G.1 provides full experimental results, extended visualizations, and additional analyses, including ablations over the number of supervised trajectories (Appendix G.2) and motion vocabulary size (Appendix G.3). It also includes behavioral cloning and action decoder convergence across multiple supervision budgets, where the same qualitative trend observed in the main text consistently holds. The supervision ablation further shows that OTF-LAM-Dino makes more effective use of additional action supervision than LAPO, particularly on \texttt{cheetah-run}, while downstream performance exhibits no clear dependence on vocabulary size within the range considered in this work.
\section{Related Works}

\paragraph{Factorized Motion and observed-transition Representations.} Recent works feature video autoencoding methods to learn decoupled representations that can separate temporal motion from spatial information, which can be used for video reconstruction and generation~\citep{ava, decovae, hivae}. 
%These works are relevant to our need for motion-centered observations because they show that video transitions can be represented through structured motion variables instead of a single dense latent, and our representation needs to capture transition variations rather than static appearances. 
However, their primary objective is video reconstruction or generation and not the decomposition of multiple simultaneously moving sources into an abstraction for action-relevant factors.
Other prior works have approached motion factorization through independent motion segmentation where the goal is to identify parts of a scene that move as either coherent or separately moving components~\citep{segmotion, eventbasedsegmotion, guesswhatmoves}. %These approaches aim to produce motion masks, region groupings, or object-centric segmentations. 
\citet{partmotion} further show that motion structure can exist below the object level through dynamic-static disentanglement to separate independently moving parts from static geometry. Our focus is not on 3D reconstruction, but instead to learn reusable primitives which capture localized visual changes.
%
%LAMs infer action-like variables from observation-only trajectories, which enable downstream tasks such as 

\paragraph{Latent Action Learning and Distractors.\\}
LAPO~\cite{lapo} introduced action-free learning for control from videos, and
LAPA~\cite{lapa} extended this idea by pretraining latent actions from
large-scale video datasets for VLA modeling. However, these methods compress
each transition into a single latent action, which becomes ambiguous in the
presence of distractors such as camera motion or other exogenous dynamics
\citep{laom, dcs}. \citet{laom} shows that distractors degrade latent action
learning and that limited action supervision improves robustness. Other
approaches instead constrain the observed transition using optical
flow~\citep{laof}, object-centric representations~\citep{oclam}, or pretrained
segmentation masks~\citep{segmentlam}, while recent work studies latent action
world models on more diverse in-the-wild videos~\citep{garrido}. In contrast, we
factorize observed transitions into multiple motion-effect factors rather than
compressing them into a single latent representation.
Several recent methods also introduce structured latent action spaces.
AC-LAM~\citep{ac-lam} represents actions with additive component vectors,
FLAM~\citep{flam} decomposes dynamics into entity-level IDM/FDM modules, and
LPWM~\citep{lpwm} models object-centric stochastic dynamics with particle-level
latents. Our approach instead factorizes transitions through patch-level
activations of a shared observed-transition codebook.

\paragraph{Compositional and Quantized Latent Representations.}
VQ-VAE introduced learned discrete codebooks as reusable latent units, and LAPA uses a VQ-VAE-style objective to learn discrete latent actions from transitions~\citep{vqvae, lapa}. Compositional VQ-VAE-style representations have also been used to decompose complex human motion into reusable discrete components, for example in speech-driven holistic 3D motion generation~\citep{yi2023generatingholistic3dhuman} and music-conditioned dance video generation~\citep{chen2025xdancerexpressivemusichuman}. Product quantization has also been used to learn compact discrete representations in visual and multimodal domains~\citep{ma2024nymeriamassivecollectionmultimodal, elnouby2023imagecompressionproductquantized}. A complementary line of work studies compositionality through object-centric representations, where objects or slots serve as the basic units of decomposition~\cite{slotattention, cjepa,dlp}. 
In contrast, we build on the idea that codebooks can provide reusable components. This provides a compositional latent interface alternative to compact latent actions, where individual codes can be analyzed through their localized effects on the reconstructed transition.

\section{Conclusion}

% We presented OTF-LAM, a bottom-up approach to latent action learning under agent ambiguity. Since observation transitions are visual effects rather than actions themselves, a monolithic latent action can entangle controllable motion with distractors, camera dynamics, and other spurious transition sources. OTF addresses this by factorizing transitions into reusable observed-transition primitives before composing them into state-aware action-like latents.
% Our motion primitives transfer across held-out carriers and morphology shifts, and the resulting latent action model, OTF-LAM, support downstream policy learning in the conventional inverse--forward dynamics framework. 
% OTF-LAM-Dino further shows that this interface can be paired with a frozen DINO representation space, improving over pixel-space prediction. These results support factorized observed effects as a useful intermediate representation for observation-only latent action learning.
% This work evaluates the approach on controlled settings and DCS distractor environments. Extending it to richer video-game domains, such as Mario-like environments with scrolling cameras, enemies, particles, and diverse non-agent dynamics, remains future work.
% More discussions and future directions are in Appendix H.

We presented OTF-LAM, a bottom-up approach to latent action learning under agent ambiguity. Rather than attempting to recover true actions directly
from observation-only transitions, OTF factorizes transitions into reusable observed-transition primitives and composes them into state-aware latent
actions that can be aligned with environment actions through downstream supervision.
The learned motion primitives transfer across held-out carriers and morphology shifts, while the resulting latent actions exhibit stronger action
alignment than conventional monolithic latent representations and support competitive downstream policy learning. OTF-LAM-Dino further demonstrates
that combining a reusable transition vocabulary with a frozen DINO representation improves over pixel-space prediction. Together, these results
suggest that reusable observed-transition vocabularies provide an effective latent action interface for observation-only learning.
This work evaluates the approach on controlled settings and DCS distractor environments. Extending it to richer video-game domains, such as Mario-like
environments with scrolling cameras, enemies, particles, and diverse
non-agent dynamics, remains future work. More discussions and future directions are provided in Appendix H.

\section*{Acknowledgement}
We thank Chen Sun, Dan Haramati, and David Klindt for helpful and meaningful discussions.

% \newpage
\bibliography{aaai2027}

\appendix

\clearpage
\onecolumn

% Reset counters
\setcounter{figure}{0}
\setcounter{table}{0}

% Prefix with appendix letter
\renewcommand{\thefigure}{A\arabic{figure}}
\renewcommand{\thetable}{A\arabic{table}}

% ------------------------------------------------------------------
% Appendix-only table of contents.
% Compile the full paper at least twice so that all \pageref values resolve.
% Section letters/numbers are included manually because the AAAI style may
% suppress rendered section numbering.
% ------------------------------------------------------------------
\begingroup
\setlength{\parindent}{0pt}
\setlength{\parskip}{0pt}

\begin{center}
{\Huge\bfseries Supplementary Material\par}
\vspace{10pt}
{\LARGE Latent Actions from Factorized Transition Effects under Agent Ambiguity\par}
\end{center}
\vspace{2.25em}

\newcommand{\appTOCsection}[3]{%
  \noindent\textbf{#1\hspace{0.45em}#2}\nobreak
  \leaders\hbox{\kern0.25em.\kern0.25em}\hfill
  \nobreak\makebox[2.2em][r]{\pageref{#3}}\par
  \vspace{0.65em}%
}
\newcommand{\appTOCsubsection}[3]{%
  \noindent\hspace{1.5em}#1\hspace{0.45em}#2\nobreak
  \leaders\hbox{\kern0.25em.\kern0.25em}\hfill
  \nobreak\makebox[2.2em][r]{\pageref{#3}}\par
  \vspace{0.4em}%
}

\appTOCsection{A.}{Problem Formulation}{app:toc:problem}
\appTOCsubsection{A.1}{Why Actions Are Not Identifiable}{app:toc:nonident}
\appTOCsubsection{A.2}{Could Anything Be Identified from Observation Transitions?}{app:toc:latent-ident}

\appTOCsection{B.}{Pretraining Details}{app:toc:pretraining}
\appTOCsubsection{B.1}{Observed Motion Signals Are Not Pure Motion}{app:toc:motion-signals}
\appTOCsubsection{B.2}{Patch Size and the Motion--Carrier Trade-off}{app:toc:patch-size}
\appTOCsubsection{B.3}{Why Spatial Reference Conditioning Encourages Transition-Centered Codes}{app:toc:conditioning}
\appTOCsubsection{B.4}{OTF Implementation Details}{app:toc:otf-impl}

\appTOCsection{C.}{OTF-LAM Implementation Details}{app:toc:otflam}
\appTOCsubsection{C.1}{Action Abstraction Module}{app:toc:action-module}
\appTOCsubsection{C.2}{OTF-LAM-Pixel}{app:toc:pixel}
\appTOCsubsection{C.3}{OTF-LAM-DINO}{app:toc:dino}
\appTOCsubsection{C.4}{Training Configuration}{app:toc:training-config}

\appTOCsection{D.}{Additional Visualizations}{app:toc:visualizations}
\appTOCsubsection{D.1}{Motion-Centered Transition Inputs}{app:toc:motion-inputs}
\appTOCsubsection{D.2}{Visualized Examples of the OTF Module}{app:toc:otf-examples}

\appTOCsection{E.}{Baselines}{app:toc:baselines}
\appTOCsection{F.}{Training Details}{app:toc:training-details}
\appTOCsubsection{F.1}{OTF-LAM-Pixel and OTF-LAM-DINO}{app:toc:training-details_1}
\appTOCsubsection{F.2}{Downstream Evaluation}{app:toc:training-details_2}

\appTOCsection{G.}{Additional Results}{app:toc:additional-results}
\appTOCsubsection{G.1}{Full Results}{app:toc:full-results}
\appTOCsubsection{G.2}{Effect of Supervision Amount}{app:toc:supervision-amount}
\appTOCsubsection{G.3}{Effect of Motion Vocabulary Size}{app:toc:vocabulary-size}

\appTOCsection{H.}{Discussion}{app:toc:discussion}

\endgroup
\clearpage

\section{\huge A. Problem Formulation}
\label{app:toc:problem}

\subsection{A.1 Why Actions Are Not Identifiable}
\label{app:toc:nonident}

The issue can be understood from a simple many-to-one function.
Consider
\[
f(u)=u^2.
\]
Since
$f(1)=f(-1)=1$,
observing the output \(1\) does not uniquely determine whether the input
was \(1\) or \(-1\). The same ambiguity arises in our setting whenever
distinct latent causes produce the same observable transition.
For the rendering map, distinct physical state changes may produce the
same visual transition:
\[
R_{\mathcal E}(s_t,\Delta s_t,\xi_t)
=
R_{\mathcal E}(s_t,\Delta s'_t,\xi_t),
\qquad
\Delta s_t\neq\Delta s'_t.
\]
Hence the visual transition does not uniquely determine the underlying
physical state change.
Even if the correct physical state change were known, the transition
function may still be many-to-one:
\[
F_{\mathcal E}(s_t,a_t,\xi_t)
=
F_{\mathcal E}(s_t,a'_t,\xi'_t),
\qquad
a_t\neq a'_t.
\]
Thus the same state change can be explained by different actions together
with different uncontrolled transition sources. Consequently, the true
action cannot in general be uniquely identified from observations alone.

\subsection{A.2 Could Anything Be Identified from Observation Transitions?}
\label{app:toc:latent-ident}

The non-identifiability of the true action does not necessarily imply
that no structure can be recovered from observation transitions.
A weaker question is whether the latent factors underlying a transition
could be identified up to a restricted class of transformations.

Existing identifiability results suggest that such a conclusion may be
possible under substantially stronger assumptions. Informally, suppose
that the observed transition is generated by an injective nonlinear
mixing of latent transition factors, that these factors are conditionally
independent given the current state, and that their conditional
distributions vary sufficiently across states. Under suitable regularity
and population-level estimation assumptions, related results show that
the latent factors may be recoverable up to permutation and
component-wise reparameterization
\citep{nonlinear1,nonlinear2}.

The current state may play a role analogous to the auxiliary variable in
these analyses, but identifiability requires substantially stronger
assumptions than observation-only data alone.
We do not establish such an identifiability result for OTF-LAM. Our
model is not designed to satisfy the assumptions underlying existing
nonlinear ICA theory, and we therefore make no claim that OTF-LAM
recovers the true latent transition factors, even up to permutation or
component-wise transformations. 

% 그럼 우리는 무엇을 하는가
Then what OTF-LAM does is imposing a discrete and compositional structure on the
transition representation. During pretraining, local transition features
are mapped to a finite set of discrete codes, providing a shared vocabulary
for representing observed transitions. This structure does not uniquely
identify the underlying physical factors: code indices remain arbitrary
up to permutation, and individual codes need not correspond one-to-one
with true actions or other generative factors. The resulting codes should
therefore be viewed as learned transition primitives whose action semantics
are determined only through downstream supervision.

\section{\huge B. Pretraining Details}
\label{app:toc:pretraining}

\subsection{B.1 Observed Motion Signals Are Not Pure Motion}
\label{app:toc:motion-signals}
\label{app:observed_motion_not_pure}

The motion input used in this work is computed from image observations,
for example through first- or second-order temporal differencing.
Although such differencing suppresses static appearance, it does not
directly expose physical motion. A pixel-level velocity or acceleration
signal remains an observed effect of motion mediated by the local visual
content of the scene.

More formally, let $m_{t,p} \in \mathcal{M}$ denote an underlying local
motion effect at patch $p$, and let $\xi_{t,p} \in \Xi$ denote the local
visual carrier through which that effect becomes observable. The measured
signal can be written as
\begin{equation}
    y_{t,p} = \Psi(m_{t,p}, \xi_{t,p}),
\end{equation}
for an image-formation-dependent map $\Psi$. The pretraining input
therefore lies in the space of observed motion effects
\begin{equation}
    \mathcal{Y}
    =
    \{\Psi(m,\xi) : m \in \mathcal{M}, \xi \in \Xi\},
\end{equation}
rather than in the space of physical motions $\mathcal{M}$ alone.
The reference-frame conditioning branch is intended to reduce, which is explained with more details in Appendix B.3.

\subsection{B.2 Patch Size and the Motion--Carrier Trade-off}
\label{app:toc:patch-size}
\label{app:patch_size_tradeoff}

Patch size controls the granularity at which observed transition effects are quantized. 
It therefore determines an important trade-off between motion locality and carrier ambiguity.

If patches are too small, each token contains only a limited visual neighborhood. 
This reduces leakage from object identity or large-scale appearance, but it can make local motion difficult to infer. 
Small patches are more sensitive to noise, texture aliasing, and the aperture problem~\citep{detopt,iterative}: a short edge fragment may not contain enough information to distinguish translation, rotation, or deformation. 
In such cases, the codebook may learn overly local residual patterns rather than stable motion-effect primitives.

If patches are too large, each token contains richer spatial context and may provide more reliable evidence for local motion. 
However, larger patches are also more likely to contain multiple motion sources, object-part identity, texture statistics, or foreground-background boundaries. 
As a result, the learned codes may become more appearance- or part-specific, approaching object-conditioned transition templates rather than reusable observed-transition primitives. 
In this regime, the effective code space may expand from a compact motion-effect basis toward a larger product space of motion and local carrier factors.

The desired patch scale is therefore task-dependent. 
A useful patch should be large enough to contain sufficient local structure for estimating a dominant transition effect, but small enough to avoid mixing multiple independent motion sources or encoding full object identity. 
We set the motion-token resolution to match the $14 \times 14$ patch grid
used by DINO, preserving a direct spatial correspondence between the
motion and visual representations. We do not vary the patch size in this
work, so its effect on the trade-off between locality and visual-carrier
content remains unexamined.

\subsection{B.3 Why Spatial Reference Conditioning Encourages Transition-Centered Codes}
\label{app:toc:conditioning}
\label{app:why_otf_conditioning}

The OTF decoder receives both the quantized transition representation and
a spatial feature map extracted from the current frame.
Let
\[
R=s_\gamma(X)
\]
denote the spatial reference feature map and
\[
Z=q_\phi(O)
\]
the quantized representation of the motion-space target \(O\).
The decoder reconstructs
\[
\hat O
=
D_\eta(Z,R).
\]
We assume that the target transition can be locally decomposed as
\[
O_i
=
F(M_i,R_{\mathcal N(i)})
+
\varepsilon_i,
\]
where \(M_i\) denotes the local transition effect,
\(R_{\mathcal N(i)}\) the reference features within the receptive field,
and \(\varepsilon_i\) is zero-mean reconstruction noise.

\paragraph{Observation.}

The decoder has access to the spatially aligned reference feature
\(R\).
Then any component of the reconstruction target that is predictable from
\(R\) need not be represented in the discrete code \(Z\).
Consequently, minimizing reconstruction error places less pressure on the
quantized representation to encode appearance, texture, object identity,
or local geometry already contained in the reference branch.
Instead, the bottleneck capacity is preferentially allocated to the
remaining transition information.
Formally, conditioning can only decrease the uncertainty that must be
resolved through the bottleneck,
\[
H(O\mid R)
\le
H(O),
\]
so the decoder requires no more information from \(Z\) than in the
unconditioned case.
Spatial alignment is also important.
Because each transition token is decoded together with the corresponding
reference neighborhood, the decoder can combine

\[
\underbrace{M_i}_{\text{transition}}
\qquad\text{and}\qquad
\underbrace{R_{\mathcal N(i)}}_{\text{appearance}}
\]

locally.
As a result, the same transition primitive may be reused across different
objects, textures, or environments without requiring separate codes for
each visual carrier.
However, the motion-space target may still contain appearance-correlated cues, and
a finite-capacity encoder may encode such information whenever it reduces
the reconstruction loss.
Instead, spatial reference conditioning changes the optimization pressure:
appearance information is available outside the discrete bottleneck,
making transition-centered codes a more efficient solution.

\subsection{B.4 OTF Implementation Details}
\label{app:toc:otf-impl}
\label{app:stage1_details}

\paragraph{Encoder.}
The input is a motion observation $o_t \in \mathbb{R}^{C \times H \times W}$.
The encoder partitions $o_t$ into non-overlapping spatial patches and embeds each patch independently using a lightweight MLP patch encoder. 
We concatenate normalized spatial coordinates to each flattened patch representation before the MLP, so that the encoder has access to coarse patch location. 
The output is a grid of latent patch embeddings $\{f_{t,i}\}_{i=1}^{P}$, where each embedding corresponds to a local transition region.

\paragraph{Local code assignment.}
Each patch embedding is discretized using a learned codebook $\mathcal{C}=\{c^{(1)},\dots,c^{(K)}\}$, where patch $i$ is assigned to its nearest codebook entry. 
To stabilize training, we use a short warmup phase in which quantization is disabled and the model behaves as a continuous autoencoder. 
After warmup, the codebook is initialized with $k$-means clustering over encoder patch embeddings collected from the training data, and discrete quantization is enabled.

\paragraph{Decoder input construction.}
For each code $k$, we compute an occupancy map $M_t^{(k)}$ indicating which patches are assigned to the code, and a normalized usage weight $w_t^{(k)}$ equal to the fraction of assigned patches. 
Together, these define the observed-transition factor $e_{t,k}=(c^{(k)},M_t^{(k)},w_t^{(k)})$. Each factor is mapped to a descriptor $h_{t,k}=\rho_\eta(e_{t,k})$ using a shared MLP. 
The descriptors are then placed back onto the patch grid using the occupancy maps
\[
H_t(u,v)=\sum_{k=1}^{K}M_t^{(k)}(u,v)h_{t,k}.
\]

\paragraph{Decoder.}
The decoder reconstructs the motion observation from the spatial feature map $H_t$. 
It uses a convolutional decoding network with bilinear upsampling to produce an output at the original resolution. 
A separate convolutional encoder extracts spatial context features from the current frame $x_t$, and these features are combined with $H_t$ before decoding. % 이것도 dino 으면 좋았을뻔
The reference pathway provides appearance and support information, while the codebook pathway provides the factorized transition information.

\paragraph{Training.}
The objective combines a reconstruction loss with the standard VQ-VAE~\cite{vqvae} codebook, commitment, and orthogonal losses
\begin{equation}
    \mathcal{L}
=
\mathcal{L}_{\mathrm{rec}}
+
\lambda_{\mathrm{code}}\mathcal{L}_{\mathrm{code}}
+
\lambda_{\mathrm{commit}}\mathcal{L}_{\mathrm{commit}}
+
\lambda_{\mathrm{orth}}\mathcal{L}_{\mathrm{orth}}.
\end{equation}

The codebook is updated by exponential moving average updates, where orthogonality primarily serves as a diagnostic for codebook diversity. To reduce code collapse, rarely used codebook entries are periodically reinitialized using randomly sampled encoder embeddings. 
We train with AdamW \citep{loshchilov2018decoupled} and gradient clipping. All OTF tokenizers are trained for $10$ epochs, with batch size 512, learning rate 1e-4, codebook vector dimension 32. 
For all DCS~\cite{dcs} environment, we use gradient transform to each frame and subtract each other to construct motion input. Each model is being trained in a single L40 GPU.

% All FiLM layers use the residual affine form
% \[
%     \operatorname{FiLM}(a;c) = (1+\gamma(c)) \odot a + \beta(c).
% \]
% For vector features, $\gamma$ and $\beta$ are linear projections from the
% condition vector. For spatial feature maps, the projections output per-channel
% scalars and broadcast them over spatial dimensions. The optional spatial decoder
% FiLM variant instead predicts per-pixel $\gamma$ and $\beta$ maps with a
% $1\times 1$ convolutional head.

\section{\huge C. OTF-LAM Implementation Details}
\label{app:toc:otflam}
\label{app:otf_lam_details}

We describe the two instantiations of OTF-LAM used in our experiments:
OTF-LAM-Pixel, which predicts future observations in pixel space, and
OTF-LAM-DINO, which predicts future representations from a frozen DINOv2
encoder. Both variants use the same pretrained and frozen OTF module and
share the same action abstraction architecture. They differ in the source
of the current-state features and in the forward prediction target.

Let \(B\) denote the batch size, \(N=hw\) the number of spatial patches,
and \(K\) the size of the OTF codebook. In our experiments, the OTF patch
grid is \(h\times w=14\times14\), giving \(N=196\) local transition
assignments. The OTF encoder and codebook remain frozen throughout
OTF-LAM training.

\subsection{C.1 Action Abstraction Module}
\label{app:toc:action-module}
\label{app:action_abstraction_details}

For each spatial location \(i\), the frozen OTF module provides the
quantized transition code assigned to that patch. We denote its code
embedding by
\[
c_{t,i}\in\mathbb{R}^{d_c},
\qquad i=1,\ldots,N.
\]
The current observation is separately encoded into spatial state features
\[
s_{t,i}\in\mathbb{R}^{d_s}.
\]
The origin and dimension of \(s_{t,i}\) depend on the model instantiation
and are described below.

We construct a state-aware transition token at each patch by combining
the state feature, the assigned OTF code, and a learnable positional
embedding:
\[
u_{t,i}
=
\phi_{\mathrm{in}}
\left(
[s_{t,i},c_{t,i},p_i]
\right),
\qquad
U_t=\{u_{t,i}\}_{i=1}^{N},
\]
where \(p_i\in\mathbb{R}^{d_a}\) is a learnable positional embedding and
\(\phi_{\mathrm{in}}\) consists of LayerNorm followed by a linear
projection into the action-aggregator dimension \(d_a\).

A fixed global pooling operation would summarize all patch tokens using
the same aggregation rule. We instead introduce \(L\) learnable action
queries
\[
Q=\{q_\ell\}_{\ell=1}^{L},
\qquad
Q\in\mathbb{R}^{L\times d_a}.
\]
The queries aggregate the patch-level transition tokens through
cross-attention:
\[
H_t
=
Q+
\operatorname{CrossAttn}
\left(
\operatorname{LN}(Q),
\operatorname{LN}(U_t),
\operatorname{LN}(U_t)
\right).
\]
The resulting action tokens are further refined through self-attention,
allowing information aggregated by different queries to be integrated
before forming the final latent action:
\[
z_t^{\mathrm{act}}
=
\phi_{\mathrm{out}}
\left(
\operatorname{Flatten}
\left(
\operatorname{Transformer}_{\mathrm{act}}(H_t)
\right)
\right).
\]

In all experiments, we use \(L=4\) action queries, aggregator dimension
\(d_a=256\), four cross-attention heads, and a depth-two Transformer with
MLP dimension \(1024\) and dropout \(0.1\). The output projection is
\[
\phi_{\mathrm{out}}
=
\operatorname{Linear}
\circ
\operatorname{LayerNorm},
\]
mapping the flattened four action tokens to
\(z_t^{\mathrm{act}}\in\mathbb{R}^{256}\).

\subsection{C.2 OTF-LAM-Pixel}
\label{app:toc:pixel}
\label{app:otf_lam_pixel}

OTF-LAM-Pixel uses a learned convolutional state encoder and predicts the
future RGB observation.

\paragraph{Pixel state encoder.}

Given the current observation
\[
x_t\in\mathbb{R}^{B\times C\times H\times W},
\]
we first compute the OTF code-usage vector
\[
w_t\in\mathbb{R}^{B\times K},
\]
where each entry records the fraction of patches assigned to the
corresponding code. The usage vector is projected into a conditioning
vector
\[
c_t^{\mathrm{occ}}
=
W_{\mathrm{occ}}w_t.
\]

The current frame is processed by a convolutional encoder with
strided downsampling and channel dimensions \(32\), \(64\), and \(128\).
The projected usage vector modulates each encoder stage through FiLM
conditioning. The resulting feature map is aligned with the OTF patch grid
and written as
\[
S_t^{\mathrm{pix}}
=
\{s_{t,i}^{\mathrm{pix}}\}_{i=1}^{N},
\qquad
s_{t,i}^{\mathrm{pix}}\in\mathbb{R}^{128}.
\]

For the Pixel instantiation, each state-aware transition token is
constructed as
\[
u_{t,i}^{\mathrm{pix}}
=
W_{\mathrm{pix}}
\operatorname{LN}
\left(
[s_{t,i}^{\mathrm{pix}},c_{t,i},p_i]
\right)
\in\mathbb{R}^{256},
\]
where
\[
s_{t,i}^{\mathrm{pix}}\in\mathbb{R}^{128},
\qquad
c_{t,i}\in\mathbb{R}^{32},
\qquad
p_i\in\mathbb{R}^{256}.
\]

\paragraph{Pixel-space forward dynamics.}

The forward predictor receives the current-state tokens and the latent
action. The \(128\)-dimensional state tokens are first projected into the
predictor dimension:
\[
X_t^{\mathrm{pix}}
=
W_s
\operatorname{LN}
\left(
S_t^{\mathrm{pix}}
\right)
\in
\mathbb{R}^{B\times N\times d_p},
\]
where \(d_p=384\).

The latent action is separately projected into a global conditioning token
\[
g_t
=
W_g z_t^{\mathrm{act}}
\in\mathbb{R}^{B\times d_p},
\]
which is prepended to the spatial state tokens. Learned predictor
positional embeddings are then added to the resulting \(N+1\) tokens.
A depth-two Transformer predictor processes these tokens:
\[
Y_t
=
P_\theta
\left(
[g_t;X_t^{\mathrm{pix}}],
z_t^{\mathrm{act}}
\right).
\]
The latent action is additionally projected and added to the spatial tokens
at each Transformer layer.

After removing the global token, the remaining \(N=196\) tokens are
reshaped into a \(14\times14\) spatial feature map. A \(1\times1\)
convolution reduces the channel dimension, and a convolutional RGB head
progressively upsamples the feature map to the original observation
resolution. The model predicts a residual update
\[
\Delta x_t
=
D_{\mathrm{rgb}}
\left(
Y_t^{\mathrm{spatial}}
\right),
\qquad
\hat{x}_{t+\tau}
=
x_t+\Delta x_t.
\]

The predictor uses dimension \(384\), depth \(2\), six attention heads,
MLP dimension \(1536\), and dropout \(0.1\).

\paragraph{Training objective.}

OTF-LAM-Pixel is trained using future-frame prediction:
\[
\mathcal{L}_{\mathrm{OTF\text{-}LAM\text{-}Pixel}}
=
\left\|
\hat{x}_{t+\tau}
-
x_{t+\tau}
\right\|_2^2.
\]
The pretrained OTF encoder and codebook are frozen. Gradients are applied
only to the pixel state encoder, action abstraction module, and forward
prediction model.

\subsection{C.3 OTF-LAM-DINO}
\label{app:toc:dino}
\label{app:otf_lam_dino}

OTF-LAM-DINO replaces pixel-space prediction with prediction in the
representation space of a frozen DINOv2 encoder. It uses the same frozen
OTF module and the same action abstraction architecture as
OTF-LAM-Pixel.

\paragraph{Frozen visual state.}

Given an observation \(x_t\), we obtain the current and future visual
representations using a frozen DINOv2 encoder \(D\):
\[
S_t^{\mathrm{dino}}
=
D(x_t),
\qquad
S_{t+\tau}^{\mathrm{dino}}
=
\operatorname{sg}
\left(
D(x_{t+\tau})
\right),
\]
where \(\operatorname{sg}(\cdot)\) denotes stop-gradient.

We use the frozen \texttt{facebook/dinov2-small} model. Input observations
are resized to \(196\times196\), producing a \(14\times14\) patch grid that
is spatially aligned with the OTF grid. The resulting patch-token
representation is
\[
S_t^{\mathrm{dino}}
=
\{s_{t,i}^{\mathrm{dino}}\}_{i=1}^{N},
\qquad
s_{t,i}^{\mathrm{dino}}
\in\mathbb{R}^{384}.
\]

For the DINO instantiation, the action-abstraction input tokens are
\[
u_{t,i}^{\mathrm{dino}}
=
W_{\mathrm{dino}}
\operatorname{LN}
\left(
[s_{t,i}^{\mathrm{dino}},c_{t,i},p_i]
\right)
\in\mathbb{R}^{256},
\]
where
\[
s_{t,i}^{\mathrm{dino}}\in\mathbb{R}^{384},
\qquad
c_{t,i}\in\mathbb{R}^{32},
\qquad
p_i\in\mathbb{R}^{256}.
\]

\paragraph{Feature-space forward dynamics.}

The forward predictor estimates the future DINO patch-token
representation:
\[
\widehat{S}_{t+\tau}^{\mathrm{dino}}
=
P_\theta
\left(
S_t^{\mathrm{dino}},
z_t^{\mathrm{act}}
\right).
\]

The current DINO tokens are normalized and projected into the predictor
dimension:
\[
X_t^{\mathrm{dino}}
=
W_x
\operatorname{LN}
\left(
S_t^{\mathrm{dino}}
\right)
\in
\mathbb{R}^{B\times N\times384}.
\]
The latent action is projected into the same dimension and used as global
conditioning for the predictor. In addition to its initial injection, a
separate projection of \(z_t^{\mathrm{act}}\) is added to the spatial tokens
at each Transformer layer. Local OTF transition codes are used to construct
the latent action but are not directly provided to the forward predictor.

The DINO predictor uses dimension \(384\), depth \(2\), six attention
heads, MLP dimension \(1536\), and dropout \(0.1\).

\paragraph{Training objective.}

The model is trained with a feature-prediction loss in the frozen DINO
representation space:
\[
\mathcal{L}_{\mathrm{OTF\text{-}LAM\text{-}DINO}}
=
\left\|
\widehat{S}_{t+\tau}^{\mathrm{dino}}
-
S_{t+\tau}^{\mathrm{dino}}
\right\|_2^2.
\]
The OTF module and DINOv2 encoder remain frozen throughout training.
Gradients are applied only to the action abstraction module and the
feature-space forward predictor.

\subsection{C.4 Training Configuration}
\label{app:toc:training-config}
\label{app:otf_lam_training}

Both variants are trained for \(20{,}000\) optimization steps with batch
size \(512\), AdamW, learning rate \(10^{-4}\), zero weight decay, and
gradient clipping with maximum norm \(1.0\). The OTF codebook size is
selected from
$K\in\{16,32,64,128\}$, where each models are trained with a single H100 GPU.

% TODO: Add any dataset-specific preprocessing or model-selection details
% that are not already reported in the experimental setup.

% sth like default params
\section{\huge D. Additional Visualizations}
\label{app:toc:visualizations}
\label{app:additional_visualizations}

\subsection{D.1 Motion-Centered Transition Inputs}
\label{app:toc:motion-inputs}
\label{app:motion}

To learn reusable observed-transition primitives effectively, the input should reduce semantic appearance information present in RGB frames while preserving the visual evidence of motion. 
We therefore consider several simple motion-centered input spaces. 
Specifically, we construct motion inputs by combining two temporal difference operators, which define the order of motion, with three frame-level transforms, which change the visual carrier through which motion is represented. 
Let $\phi$ denote an image-gradient transformation applied independently to each frame.
For first-order velocity inputs, we define
\begin{equation}
o_t^{\mathrm{vel},\phi}
=
\phi(x_{t+\tau})-\phi(x_t).
\end{equation}
For second-order acceleration inputs, we define
\begin{equation}
o_t^{\mathrm{acc},\phi}
=
\phi(x_{t+\tau}) - 2\phi(x_t) + \phi(x_{t-\tau}).
\end{equation}
In all cases, the transform is applied before temporal differencing.

\subsection{D.2 Visualized Examples of the OTF Module}
\label{app:toc:otf-examples}
\label{app:otfvis}

\begin{figure*}[h!]
\centering
\includegraphics[width=\textwidth]{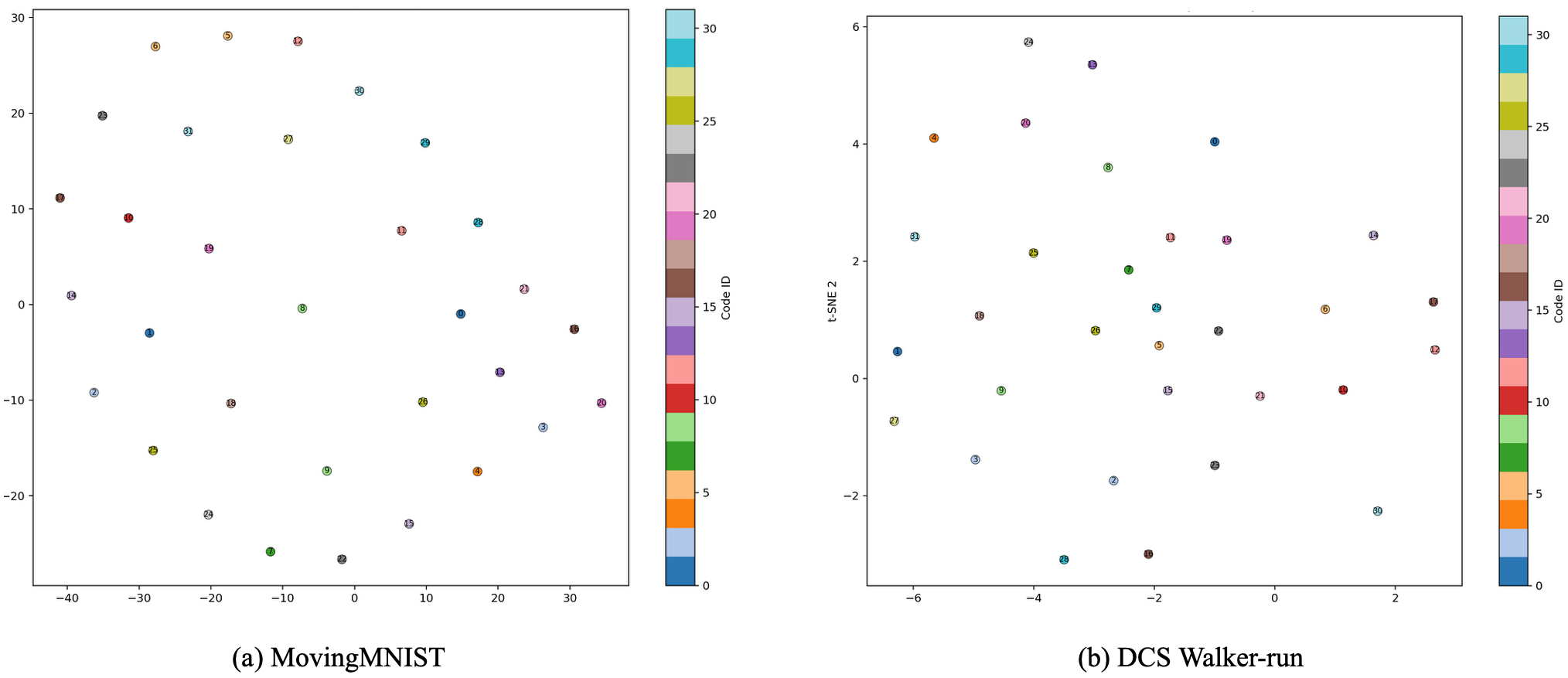}
\caption{
Visualization of the learned codebook From OTF. 
Each point denotes one learned codebook vector, and the two-dimensional coordinates are obtained by applying t-SNE~\cite{tsne} to the code embeddings. 
}
\label{fig:codebook}
\end{figure*}
We provide additional qualitative visualizations of the learned observed-transition vocabulary. 
This visualization provides a qualitative view of how the discrete observed-transition primitives are organized in the learned vocabulary.

\paragraph{Code Assignment.} 
\begin{figure*}[h!]
\centering
\includegraphics[width=0.6\textwidth]{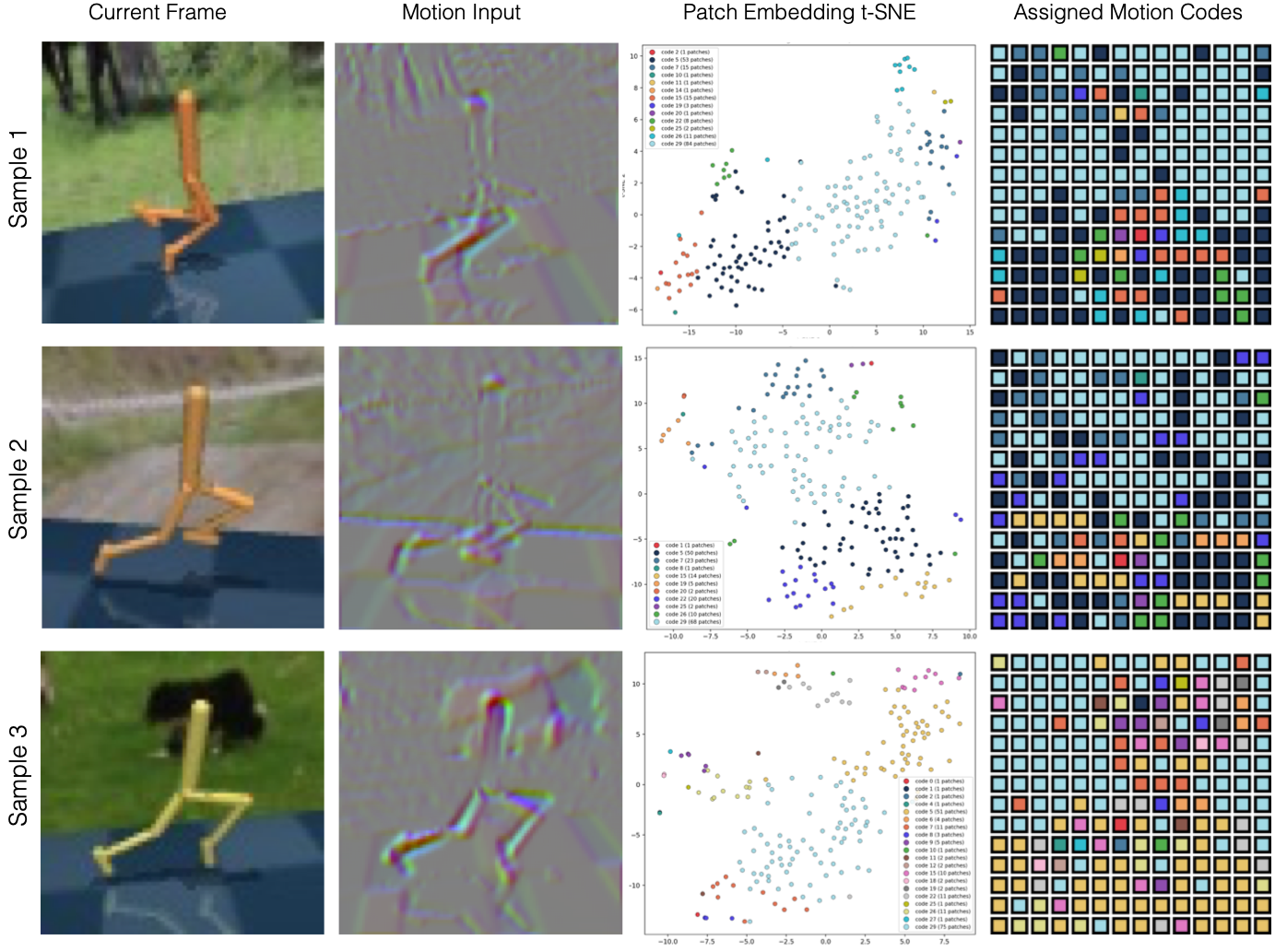}
\caption{
Example code assignments for walker-run transitions from Distracting Control Suite. 
}
\label{fig:code_eg2}
\end{figure*}
\begin{figure*}[h!]
\centering
\includegraphics[width=0.6\textwidth]{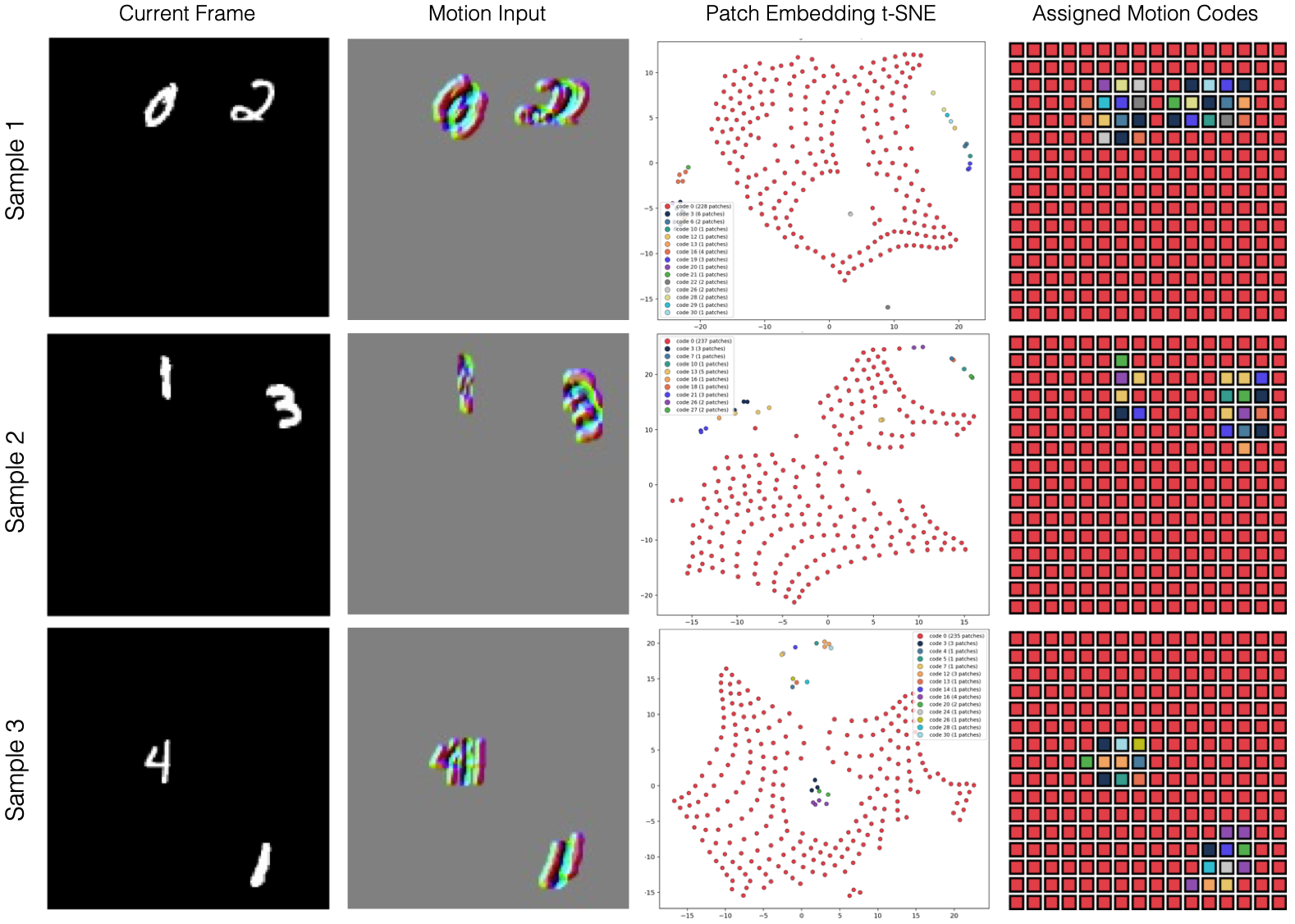}
\caption{
Example code assignments for controlled Moving MNIST transitions. 
}
\label{fig:code_eg}
\end{figure*}
Fig.~\ref{fig:code_eg2} shows example assignments on individual \texttt{walker-run} transitions. Similarly, Fig.~\ref{fig:code_eg} shows example assignments on individual Moving MNIST samples. 
For each sample, we visualize the current frame, the motion-centered input, the t-SNE~\cite{tsne} projection of patch embeddings, and the corresponding assigned codebook entries on the patch grid or assigned observed-transition primitives on the patch grid. 
These examples illustrate how local transition patterns are mapped to discrete codebook primitives across different spatial regions.

\begin{figure*}[h]
\centering
\includegraphics[width=0.6\textwidth]{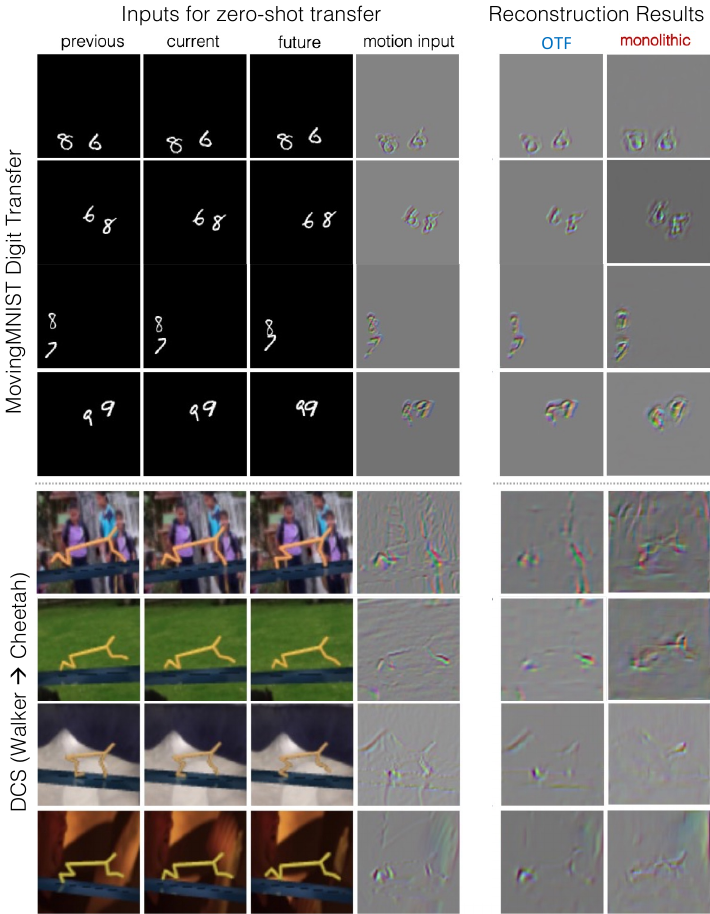}
\caption{Extra visualizations from zero-shot transfer experiments.}
\label{fig:zeroshot_app}
\end{figure*}

\paragraph{More visualization from Zero-shot Transferability} 
Figure~\ref{fig:zeroshot_app} shows extra visualizations for both MovingMNIST and DCS zero-shot transfer experiments.

\section{\huge E. Baselines}
\label{app:toc:baselines}
\label{app:baseline_details}

\subsection{E.1 LAPO~\cite{lapo}}
LAPO is a monolithic latent action model that learns action-like
representations from observation-only trajectories.
Its inverse dynamics model (IDM) infers a latent action from consecutive
observations, while its forward dynamics model (FDM) reconstructs or predicts
the subsequent observation conditioned on the current observation and the
inferred latent action.
The entire scene transition is represented using a single latent action,
providing a standard monolithic baseline for comparison with our localized
and factorized representation.

\paragraph{Training Details}
We train LAPO for $20$k optimization steps following the experimental
configuration used by LAOM~\cite{laom}.
We evaluate latent action dimensions of $128$ and $256$.
Unless otherwise specified, the remaining settings follow the 
implementation from LAOM~\cite{laom}.

\subsection{E.2 FLAM~\cite{flam}}
FLAM extends monolithic latent action models by
decomposing a scene into multiple independent factors.
Each factor infers its own latent action and predicts its own next-step factor
representation, allowing the model to capture transitions involving multiple
independently changing entities.
In contrast to a single global latent action, this factorized formulation is
designed to model complex multi-entity dynamics through a collection of
factor-specific latent actions. We have additional discussion comparing object centric methods and motion centric methods (ours) in Appendix H.

\paragraph{Training Details}
We follow the authors' implementation.
We use $d_{\mathrm{model}}=256$, two transformer blocks, and $128$ latent
$z$ channels.
We evaluate models with $4$ and $8$ factors, referred to as slots in the
implementation.
Each factor uses an action codebook containing $16$ codes, and the VAE--VQ KL
weight is annealed from $0$ to $2\times10^{-4}$.
We use a sub-trajectory length of $10$, a history length of $1$, and
$5$ prediction steps.
The tokenizer consists of an IMPALA encoder, a LAPO decoder, and an FSQ
quantizer.
We use an input image size of $64$, an FSQ codebook size of $1024$, levels
$(4,4,4,4,4)$, $128$ latent channels, and a codebook dimension of $128$.
We train FLAM for $10$ epochs with batch size $32$, learning rate
$10^{-4}$, zero weight decay, and gradient-norm clipping at $1.0$.
We additionally use xFormers attention, frame skip $1$, HDF5 datasets, and
random seed $42$.

\subsection{E.3 HiLAM~\cite{hilam}}
HiLAM learns temporally extended latent
skills on top of a pretrained low-level latent action model.
The pretrained model first extracts a sequence of low-level latent actions
from observation transitions, after which HiLAM aggregates their temporal
patterns into higher-level latent skills.
This hierarchical construction is intended to capture longer-horizon
structure that may not be represented by latent actions learned from
individual short-horizon transitions.

\paragraph{Training Details}
We follow the default configuration provided in the authors' implementation.
We train HiLAM for $50$k optimization steps with a learning rate of
$10^{-4}$ and a latent action dimension of $64$.

\subsection{E.4 Motion LAPO (M-LAPO)}
Motion LAPO augments LAPO with explicit motion inputs alongside the pixel
observations.
It retains LAPO's monolithic latent action formulation while exposing the
model directly to motion information.
This baseline therefore isolates the effect of explicit motion processing
from the effects of localized encoding and patchwise quantization introduced
by OTF-LAM.

\paragraph{Training Details}
We train M-LAPO for $20$k optimization steps with a latent action dimension
of $128$.
All other settings follow those used for LAPO.

\subsection{E.5 Factorized LAPO (F-LAPO)}
Factorized LAPO replaces LAPO's global latent action encoder and quantization
mechanism with an OTF-LAM-style patch encoder and patchwise quantization.
Unlike OTF-LAM, however, it does not use explicit motion inputs.
This baseline isolates the effect of learning spatially localized discrete
latent representations while retaining the remaining LAPO training
framework.

\paragraph{Training Details}
We train F-LAPO for $20$k optimization steps with a latent action dimension
of $128$.
All other settings follow those used for LAPO.

% \section{\huge F. Training Details}
% \label{app:toc:training-details}
% \label{app:training_details}

% OTF-LAM-Pixel are trained for $20$k steps, while OTF-LAM-DINO is trained for $10$k steps.

% \paragraph{Downstream}
% For downstream policy learning, the OTF-LAM and baselines (FLAM, HiLAM, LAPO, and LAPO variants) behavior cloning policy is trained for $10$ epochs, and action decoder is trained for 10k steps.
% For OTF-LAM-DINO, policy is trained for $5$ epochs, and 5k steps for action decoder training

\section{\huge F. Training Details}
\label{app:toc:training-details}
\label{app:training_details}

\subsection{F.1 OTF-LAM-Pixel amd OTF-LAM-DINO}
\label{app:toc:training-details_1}

We train separate models for each environment and codebook size
$K\in\{16,32,64,128\}$.
All models are trained for $20$k optimization steps with batch size $512$
using AdamW (learning rate $10^{-4}$, weight decay $0$) and gradient norm
clipping of $1.0$.
The pretrained motion encoder and codebook are frozen throughout training.

\subsection{F.2 Downstream Evaluation}
\label{app:toc:training-details_2}

We evaluate all pretrained latent action models using a two-stage downstream
pipeline consisting of behavior cloning followed by supervised action
decoding.

\paragraph{Behavior Cloning}

The behavior-cloning policy receives a stack of three RGB frames and predicts
the latent action inferred by the frozen latent action model.
We use the first $1{,}000$ trajectories of each environment, batch size
$512$, AdamW with learning rate $10^{-4}$ and zero weight decay, and random
seeds $\{0,1,2\}$.
OTF-LAM-Pixel policies are trained for $10$ epochs, while
OTF-LAM-DINO policies are trained for $5$ epochs.

\paragraph{Action Decoding}
The action decoder maps the predicted latent action together with the policy
state representation to the environment action.
We evaluate supervision budgets of
$\{16,32,64,128\}$ trajectories, where we use
$\{2.5$k, $5$k, $10$k, $20$k$\}$ optimization steps for each.
All decoders are trained with batch size $512$ using AdamW with learning
rate $3\times10^{-4}$ and weight decay $0$.
Each trained policy is evaluated for $10$ episodes. We also train for three seeds, matching with behavior cloning.

\section{\huge G. Additional Results}
\label{app:toc:additional-results}

\subsection{G.1 Full Results}
\label{app:toc:full-results}

\begin{table}[h]
\centering
  {\centering
   \tabcolsep=0.7cm
   \renewcommand{\arraystretch}{1}
   \small
   \centering
   \caption{
   \centering Mean and standard deviation of average return across 10 trajectories over 3 random seeds. 
   }
   \label{tab:full_baseline}
   \begin{tabular}{@{}lcccc@{}}
   \toprule
   \multirow{2}{*}{\textbf{Model}}
   & \multicolumn{2}{c}{\textbf{Cheetah-Run}}
   & \multicolumn{2}{c}{\textbf{Walker-Run}} \\
   \cmidrule(l){2-5}
   & Mean return & Std. & Mean return & Std. \\
   \midrule
   LAPO-128~\cite{lapo}
   & 11.48 & 7.19 & 28.71 & 7.59 \\
   \midrule
   LAPO-256~\cite{lapo}
   & 13.55 & 5.40 & 23.91 & 2.82 \\
   \midrule
   M-LAPO
   & 9.68 & 4.18 & 23.93 & 4.40 \\
   \midrule
   F-LAPO
   & 11.56 & 3.39 & 23.44 & 5.79 \\
   \midrule
   FLAM-4~\cite{flam}
   & 40.52 & 27.82 & 28.44 & 6.25 \\
   \midrule
   FLAM-8~\cite{flam}
   & 26.29 & 16.90 & 29.65 & 7.28 \\
   \midrule
   LAPO-128~\cite{lapo}
   & 11.48 & 7.19 & 28.71 & 7.59 \\
   \midrule
   \textbf{OTF-LAM-Pixel} (best config.)
   & 20.48 & 19.74 & 27.45 & 7.54 \\
   \midrule
   \textbf{OTF-LAM-DINO} (best config.)
   & 43.80 & 26.98 & 28.87 & 8.23 \\
   \bottomrule
   \end{tabular}
  }%
  {}%
\end{table}%
% \usepackage{booktabs}
% \usepackage{multirow}
% \usepackage[dvipsnames]{xcolor}

\begin{table}[h!]
\centering
   \tabcolsep=0.3cm
   \renewcommand{\arraystretch}{0.9}
   \caption{Effect of supervision amount on downstream performance.}
\label{tab:supervision}
\begin{tabular}{@{}lcccccc@{}}
\toprule
\multirow{2}{*}{Model}         & \multirow{2}{*}{codebook size} & \multirow{2}{*}{supervised trajectories} & \multicolumn{2}{c}{Cheetah-Run} & \multicolumn{2}{c}{Walker-Run} \\ \cmidrule(l){4-7}
                               &                          &                                                & Mean return    & Std.    & Mean return    &  Std.   \\ \midrule
\multirow{16}{*}{OTF-LAM-DINO} & \multirow{4}{*}{16}      & 16                                             & 34.06 {\tiny\textcolor{gray}{(--)}}                                  & 19.12      & 27.68 {\tiny\textcolor{gray}{(--)}}                                  & 6.03      \\ \cmidrule(l){3-7}
                               &                          & 32                                             & 41.61 {\tiny\textcolor{ForestGreen}{($\uparrow$7.55)}}                & 22.91      & 28.74 {\tiny\textcolor{ForestGreen}{($\uparrow$1.06)}}                & 7.00      \\ \cmidrule(l){3-7}
                               &                          & 64                                             & 55.69 {\tiny\textcolor{ForestGreen}{($\uparrow$21.63)}}               & 30.49      & 32.56 {\tiny\textcolor{ForestGreen}{($\uparrow$4.88)}}                & 11.47     \\ \cmidrule(l){3-7}
                               &                          & 128                                            & 60.71 {\tiny\textcolor{ForestGreen}{($\uparrow$26.65)}}               & 30.72      & 33.91 {\tiny\textcolor{ForestGreen}{($\uparrow$6.23)}}                & 11.66     \\ \cmidrule(l){2-7}
                               & \multirow{4}{*}{32}      & 16                                             & 35.51 {\tiny\textcolor{gray}{(--)}}                                  & 27.73      & 24.00 {\tiny\textcolor{gray}{(--)}}                                  & 6.17      \\ \cmidrule(l){3-7}
                               &                          & 32                                             & 37.75 {\tiny\textcolor{ForestGreen}{($\uparrow$2.24)}}                & 22.16      & 28.08 {\tiny\textcolor{ForestGreen}{($\uparrow$4.08)}}                & 6.67      \\ \cmidrule(l){3-7}
                               &                          & 64                                             & 40.68 {\tiny\textcolor{ForestGreen}{($\uparrow$5.17)}}                & 24.50      & 30.99 {\tiny\textcolor{ForestGreen}{($\uparrow$6.99)}}                & 8.53      \\ \cmidrule(l){3-7}
                               &                          & 128                                            & 50.96 {\tiny\textcolor{ForestGreen}{($\uparrow$15.45)}}               & 28.71      & 38.30 {\tiny\textcolor{ForestGreen}{($\uparrow$14.30)}}               & 15.30     \\ \cmidrule(l){2-7}
                               & \multirow{4}{*}{64}      & 16                                             & 32.31 {\tiny\textcolor{gray}{(--)}}                                  & 22.69      & 26.73 {\tiny\textcolor{gray}{(--)}}                                  & 5.70      \\ \cmidrule(l){3-7}
                               &                          & 32                                             & 36.10 {\tiny\textcolor{ForestGreen}{($\uparrow$3.79)}}                & 21.75      & 28.36 {\tiny\textcolor{ForestGreen}{($\uparrow$1.63)}}                & 7.87      \\ \cmidrule(l){3-7}
                               &                          & 64                                             & 42.79 {\tiny\textcolor{ForestGreen}{($\uparrow$10.48)}}               & 27.81      & 33.14 {\tiny\textcolor{ForestGreen}{($\uparrow$6.41)}}                & 11.32     \\ \cmidrule(l){3-7}
                               &                          & 128                                            & 46.39 {\tiny\textcolor{ForestGreen}{($\uparrow$14.08)}}               & 31.92      & 38.97 {\tiny\textcolor{ForestGreen}{($\uparrow$12.24)}}               & 12.88     \\ \cmidrule(l){2-7}
                               & \multirow{4}{*}{128}     & 16                                             & 37.87 {\tiny\textcolor{gray}{(--)}}                                  & 22.95      & 24.98 {\tiny\textcolor{gray}{(--)}}                                  & 4.88      \\ \cmidrule(l){3-7}
                               &                          & 32                                             & 40.02 {\tiny\textcolor{ForestGreen}{($\uparrow$2.15)}}                & 24.04      & 29.04 {\tiny\textcolor{ForestGreen}{($\uparrow$4.06)}}                & 8.40      \\ \cmidrule(l){3-7}
                               &                          & 64                                             & 46.24 {\tiny\textcolor{ForestGreen}{($\uparrow$8.37)}}                & 26.94      & 30.70 {\tiny\textcolor{ForestGreen}{($\uparrow$5.72)}}                & 8.70      \\ \cmidrule(l){3-7}
                               &                          & 128                                            & 46.88 {\tiny\textcolor{ForestGreen}{($\uparrow$9.01)}}                & 33.27      & 38.54 {\tiny\textcolor{ForestGreen}{($\uparrow$13.56)}}               & 13.74     \\ \midrule
\multirow{4}{*}{LAPO-256}      & \multirow{4}{*}{-}       & 16                                             & 4.80 {\tiny\textcolor{gray}{(--)}}                                   & 1.96       & 26.20 {\tiny\textcolor{gray}{(--)}}                                  & 2.40      \\ \cmidrule(l){3-7}
                               &                          & 32                                             & 13.55 {\tiny\textcolor{ForestGreen}{($\uparrow$8.75)}}                & 5.40       & 23.91 {\tiny\textcolor{BrickRed}{($\downarrow$2.29)}}                & 2.82      \\ \cmidrule(l){3-7}
                               &                          & 64                                             & 12.62 {\tiny\textcolor{ForestGreen}{($\uparrow$7.82)}}                & 5.41       & 25.59 {\tiny\textcolor{BrickRed}{($\downarrow$0.61)}}                & 5.31      \\ \cmidrule(l){3-7}
                               &                          & 128                                            & 14.74 {\tiny\textcolor{ForestGreen}{($\uparrow$9.94)}}                & 7.32       & 25.10 {\tiny\textcolor{BrickRed}{($\downarrow$1.10)}}                & 4.48      \\ \bottomrule
\end{tabular}
\end{table}
\begin{table}[h!]
\centering
  {\centering
   \tabcolsep=0.7cm
    \renewcommand{\arraystretch}{1}
    \small
\centering
\caption{
Downstream evaluation of OTF-LAM and OTF-LAM-DINO with different motion vocabulary sizes. Std. is computed as root mean square across three random seeds.
}
\label{tab:full_ksweep}
\begin{tabular}{@{}lccccc@{}}
\toprule
\multirow{2}{*}{} & \multirow{2}{*}{K} & \multicolumn{2}{c}{\textbf{OTF-LAM}} & \multicolumn{2}{c}{\textbf{OTF-LAM-DINO}} \\ \cmidrule(l){3-6}
& & Mean return & Std. & Mean return & Std. \\ \midrule

\multirow{4}{*}{\textbf{Cheetah-Run}}
& 16  & 12.24 & 9.79  & 40.85 & 23.69 \\ \cmidrule(l){2-6}
& 32  & 17.06 & 14.66 & 43.80 & 26.98 \\ \cmidrule(l){2-6}
& 64  & 16.70 & 18.30 & 33.09 & 20.46 \\ \cmidrule(l){2-6}
& 128 & 20.48 & 19.74 & 42.92 & 26.01 \\ \midrule

\multirow{4}{*}{\textbf{Walker-Run}}
& 16  & 26.52 & 5.53 & 28.87 & 8.23 \\ \cmidrule(l){2-6}
& 32  & 25.62 & 4.67 & 27.64 & 6.57 \\ \cmidrule(l){2-6}
& 64  & 24.66 & 5.02 & 28.70 & 7.74 \\ \cmidrule(l){2-6}
& 128 & 27.45 & 7.54 & 27.65 & 8.09 \\ \bottomrule

\end{tabular}
      }%
    {}%
\end{table}%

We provide the complete quantitative results, including the full baseline
comparison (Table~\ref{tab:full_baseline}), the effect of supervision amount (Table~\ref{tab:supervision}), and the codebook-size ablation
for OTF-LAM and OTF-LAM-DINO (Table~\ref{tab:full_ksweep}).

\subsection{G.2 Effect of Supervision Amount}
\label{app:toc:supervision-amount}

Table~\ref{tab:supervision} examines the effect of increasing the amount of
supervised trajectories used for downstream policy learning. For a controlled
comparison, OTF-LAM-DINO and LAPO-256 use the same latent action dimension.
Across all codebook sizes, OTF-LAM-DINO consistently benefits from additional
supervision. Performance improves almost monotonically as the number of
labeled demonstrations increases, with the largest gains typically occurring
between the lowest and highest supervision budgets. On Cheetah-Run, the
average return often improves by roughly $10$--$30$ points depending on the
codebook size, while Walker-Run exhibits similarly consistent, although
generally smaller, improvements.

The benefit of additional supervision is particularly pronounced when using
larger codebooks. Models with larger motion vocabularies generally start from
stronger representations under limited supervision and continue to improve as
more labeled data becomes available. In contrast, smaller codebooks remain
competitive in the low-supervision regime but tend to saturate earlier.
For comparison, LAPO-256 shows substantially weaker scaling with supervision.
While increasing the number of demonstrations improves performance on
Cheetah-Run, the gains are considerably smaller than those of OTF-LAM-DINO.
On Walker-Run, additional supervision provides little benefit and occasionally
leads to slight performance degradation, suggesting that its learned latent
actions are less amenable to downstream policy learning.
Overall, these results indicate that the representations learned by
OTF-LAM-DINO remain informative across a wide range of supervision budgets,
allowing downstream policies to effectively exploit additional labeled
demonstrations, whereas conventional monolithic latent actions exhibit much
weaker scalability with supervision.

\subsection{G.3 Effect of Motion Vocabulary Size}
\label{app:toc:vocabulary-size}

We study how the size of the motion primitives affects downstream policy performance.
Table~\ref{tab:full_ksweep} reports evaluation returns for OTF-LAM and OTF-LAM-Dino with different codebook sizes $K$ on \texttt{cheetah-run} and \texttt{walker-run}, averaged over $3$ seeds, where each seed evaluates $10$ trajectories.
For decoder-based OTF-LAM, \texttt{cheetah-run} performance improves consistently as $K$ increases.
This suggests that a larger vocabulary provides a more expressive set of transition primitives for this environment.
On \texttt{walker-run}, however, performance is relatively stable across codebook sizes, with all variants achieving similar returns, indicating that the required transition structure in \texttt{walker-run} is captured even by smaller vocabularies, or that downstream control is less sensitive to additional motion primitives in this setting.
OTF-LAM-DINO shows a different pattern.
It improves over decoder-based OTF-LAM for every tested $K$ in both environments, suggesting that prediction in a frozen DINO representation space provides a useful downstream interface for the learned latent actions.
However, its performance is not monotonic in $K$.
Overall, codebook size should be viewed as an environment- and architecture-dependent capacity parameter rather than a monotonic source of improvement.
A more systematic study of how vocabulary size interacts with motion complexity, representation-space prediction, and downstream control remains future work.

\section{\huge H. Discussion}
\label{app:toc:discussion}

\subsection{Toward a General-Purpose Motion Tokenizer}
\label{app:discussion:scaling}

A promising direction for future work is to scale OTF pretraining across
heterogeneous video datasets. Similar to large-scale optical-flow training,
one could combine trajectories from different environments, embodiments, and
visual domains to learn a shared motion tokenizer rather than training a
separate tokenizer for each environment. Such a tokenizer could provide a
general vocabulary of local transition patterns that transfers across tasks
and serves as a foundation for downstream latent action learning. Developing
this type of broadly pretrained motion tokenizer will require addressing
differences in appearance, temporal scale, camera motion, and embodiment
across data sources.

\subsection{Relation to Object-Centric Learning}
\label{app:discussion:object-centric}

OTF-LAM does not require an explicit decomposition of a scene into semantic
objects. Object-centric methods based on learned slots can be sensitive to
initialization, architectural choices, and the number of slots, and their
decompositions may become unreliable in visually complex scenes. In contrast,
OTF-LAM quantizes localized transition evidence directly and composes the
resulting primitives without requiring stable object assignments. The learned
primitives need not correspond one-to-one with objects, but they provide a
simpler and potentially more robust basis for representing local dynamics.


\end{document}